\documentclass{article}

\usepackage{microtype}
\usepackage{graphicx}
\usepackage{booktabs} 

\usepackage[dvipsnames]{xcolor}

\usepackage{float}
\usepackage{scalerel}
\usepackage{pgfplots}
\usepackage{ulem}
\usepackage{multirow}
\usepackage{adjustbox}
\usepackage{colortbl}
\usepackage{fancyhdr}
\usepackage{silence}
\usepackage{numprint}
\usepackage{soul}

\usetikzlibrary{datavisualization}
\usetikzlibrary{datavisualization.formats.functions}
\usetikzlibrary{plotmarks}
\usetikzlibrary{angles,quotes,3d,math,arrows.meta,calc,positioning,fit,backgrounds,decorations.pathreplacing,calligraphy,shapes,shapes.multipart,patterns,patterns.meta}

\newcommand{\oursubsubsection}[1]{\noindent \textbf{#1}}

\newcommand{\hourglass}{\protect\scalerel*{\includegraphics{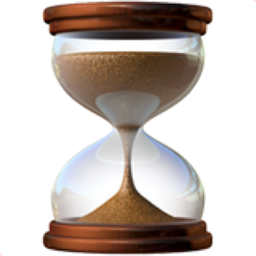}}{H}}
\newcommand{\modelname}{\mbox{\hourglass{} HDiT}}

\newcommand{\lowerisbetter}{$\downarrow$}
\newcommand{\higherisbetter}{$\uparrow$}

\setuldepth{asdf}

\let\originalleft\left
\let\originalright\right
\renewcommand{\left}{\mathopen{}\mathclose\bgroup\originalleft}
\renewcommand{\right}{\aftergroup\egroup\originalright}

\newcommand\rurl[1]{%
  \href{https://#1}{\nolinkurl{#1}}%
}

\DeclareMathAlphabet\mathbfcal{OMS}{cmsy}{b}{n}

\pgfplotsset{compat=1.18}

\newcommand{\tikzstylenodedistance}{4mm}
\newcommand{\tikzstyleinnersep}{2mm}
\newcommand{\tikzstyleminimumheight}{7mm}
\newcommand{\tikzstyleminimumwidth}{30mm}

\renewcommand{\tikzstylenodedistance}{3mm}
\renewcommand{\tikzstyleinnersep}{1.25mm}
\renewcommand{\tikzstyleminimumheight}{5.5mm}
\renewcommand{\tikzstyleminimumwidth}{25mm}

\tikzset{
    node distance=\tikzstylenodedistance
}
\tikzset{
    standard node/.style n args={1}{%
        rectangle,
        rounded corners=0.1cm,
        fill=our#1,
        draw=our#1border,
        line width=0.04cm,
        minimum height=\tikzstyleminimumheight,
        minimum width=\tikzstyleminimumwidth,
        inner sep=\tikzstyleinnersep,
        text centered,
        anchor=center,
        align=center,
    }
}
\tikzset{
    standard node circle/.style n args={1}{%
        fill=our#1,
        draw=our#1border,
        circle,
        inner sep=0.1cm,
        minimum height=0,
        minimum width=0,
    }
}
\tikzset{
    standard node circle/.prefix style = standard node
}

\tikzset{
    standard line/.style n args={0}{%
        line width=0.04cm,
        rounded corners=0.1cm,
    }
}
\tikzset{
    standard arrow/.style n args={0}{%
        -latex,
    }
}
\tikzset{
    standard arrow/.prefix style = standard line
}

\tikzset{
    simple node image/.style n args={0}{%
        rectangle,
        inner sep=0,
        text centered,
        anchor=center,
        align=center,
        node distance=0mm
    }
}

\definecolor{our}{RGB}{0,0,0}
\definecolor{ourborder}{RGB}{0,0,0}

\definecolor{ourgreen}{RGB}{46, 204, 113}
\definecolor{ourgreenborder}{RGB}{39, 174, 96}
\definecolor{ourblue}{RGB}{52, 152, 219}
\definecolor{ourblueborder}{RGB}{41, 128, 185}
\definecolor{ourorange}{RGB}{230, 126, 34}
\definecolor{ourorangeborder}{RGB}{211, 84, 0}
\definecolor{ourred}{RGB}{231, 76, 60}
\definecolor{ourredborder}{RGB}{192, 57, 43}
\definecolor{ouryellow}{RGB}{241, 196, 15}
\definecolor{ouryellowborder}{RGB}{243, 156, 18}
\definecolor{ourpurple}{RGB}{155, 89, 182}
\definecolor{ourpurpleborder}{RGB}{142, 68, 173}
\definecolor{ourturquoise}{RGB}{26, 188, 156}
\definecolor{ourturquoiseborder}{RGB}{22, 160, 133}
\definecolor{ourturquoise}{RGB}{26, 188, 156}
\definecolor{ourturquoiseborder}{RGB}{22, 160, 133}
\definecolor{ourwhite}{RGB}{236, 240, 241}
\definecolor{ourwhiteborder}{RGB}{189, 195, 199}
\definecolor{ourgray}{RGB}{149, 165, 166}
\definecolor{ourgrayborder}{RGB}{127, 140, 141}

\usepackage{hyperref}

\usepackage[accepted]{icml2024}

\usepackage{amsmath}
\usepackage{amssymb}
\usepackage{mathtools}
\usepackage{amsthm}

\usepackage{caption}
\usepackage{subcaption}

\usepackage{array}

\usepackage{silence}  
\usepackage{etoolbox}

\usepackage[capitalize,noabbrev]{cleveref}

\theoremstyle{plain}

\theoremstyle{definition}

\theoremstyle{remark}

\icmltitlerunning{Hourglass Diffusion Transformers}

\begin{document}

\twocolumn[
\icmltitle{Scalable High-Resolution Pixel-Space Image Synthesis with\\ Hourglass Diffusion Transformers}



\icmlsetsymbol{equal}{*}

\begin{icmlauthorlist}
\icmlauthor{Katherine Crowson}{equal,stability}
\icmlauthor{Stefan Andreas Baumann}{equal,lmu}
\icmlauthor{Alex Birch}{equal,alex}
\icmlauthor{Tanishq Mathew Abraham}{stability}
\icmlauthor{Daniel Z. Kaplan}{daniel}
\icmlauthor{Enrico Shippole}{enrico}
\end{icmlauthorlist}

\icmlaffiliation{stability}{Stability AI, United States}
\icmlaffiliation{lmu}{CompVis @ LMU Munich, Germany}
\icmlaffiliation{alex}{Birchlabs, England, United Kingdom}
\icmlaffiliation{daniel}{realiz.ai, New York, United States}
\icmlaffiliation{enrico}{Independent Researcher, Florida, United States}

\icmlcorrespondingauthor{Katherine Crowson}{crowsonkb@gmail.com}
\icmlcorrespondingauthor{Stefan Baumann}{\mbox{stefan.baumann@lmu.de}}
\icmlcorrespondingauthor{Alex Birch}{alex@birchlabs.co.uk}

\icmlkeywords{Diffusion Models,Generative Models,High-resolution Image Synthesis}

\vskip 0.3in
]



\printAffiliationsAndNotice{\icmlEqualContribution} 


\setlength{\fboxrule}{0pt}
\begin{abstract}
We present the Hourglass Diffusion Transformer (HDiT), an image-generative model that exhibits linear scaling with pixel count, supporting training at high resolution (e.g. $1024 \times 1024$) directly in pixel-space.
Building on the Transformer architecture, which is known to scale to billions of parameters, it bridges the gap between the efficiency of convolutional U-Nets and the scalability of Transformers.
HDiT trains successfully without typical high-resolution training techniques such as multiscale architectures, latent autoencoders or self-conditioning.
We demonstrate that HDiT performs competitively with existing models on ImageNet $256^2$, and sets a new state-of-the-art for diffusion models on FFHQ-$1024^2$.
Code is available at\\\mbox{\rurl{github.com/crowsonkb/k-diffusion}}.

\end{abstract}

\section{Introduction}\label{sec:intro}
\begin{figure}[htb]
    \centering
    \includegraphics[width=\columnwidth]{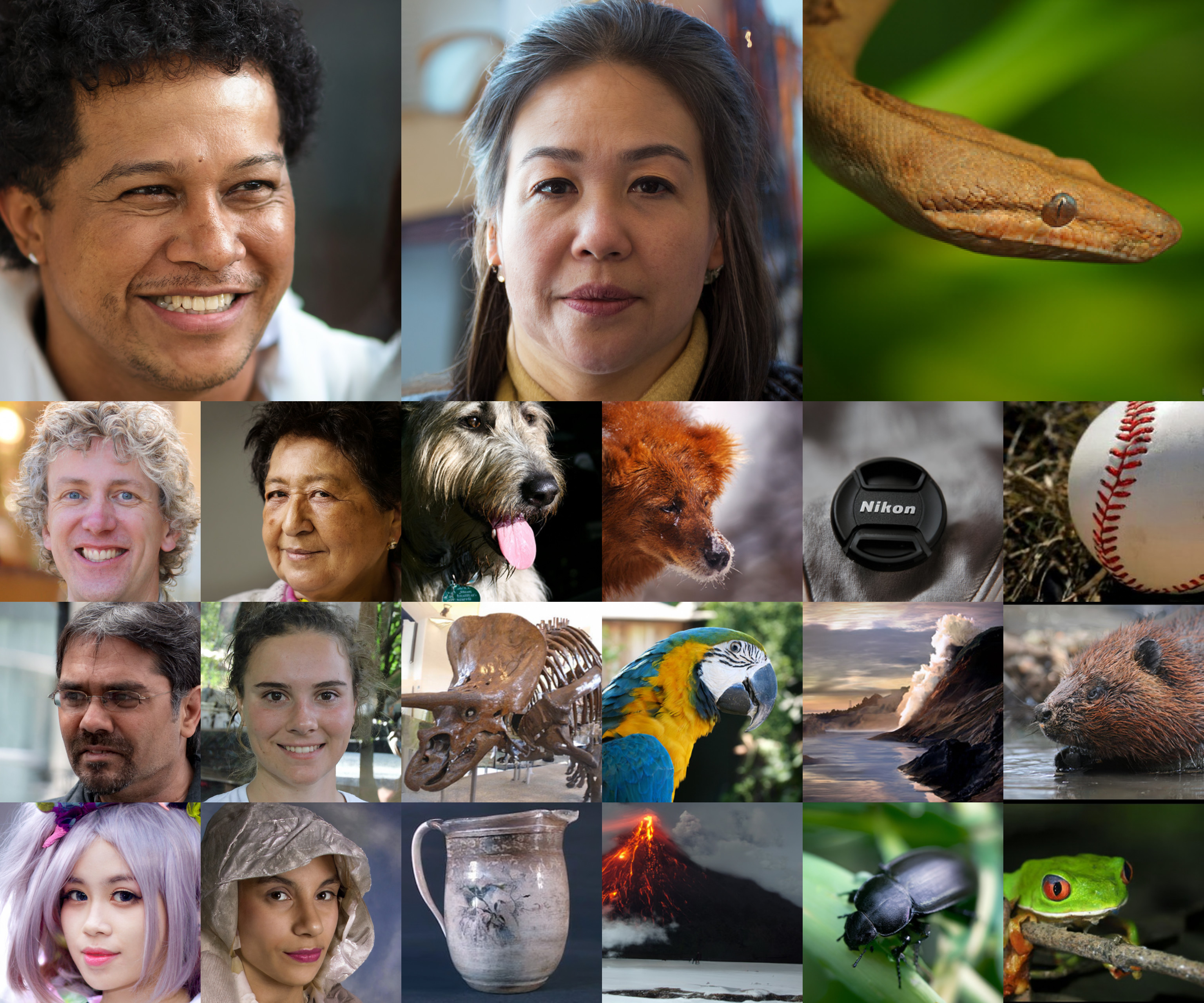}
    \caption{Samples generated directly in RGB pixel space using our \modelname{} models trained on FFHQ-$1024^2$ and ImageNet-$256^2$.}
    \label{fig:teaser}
\end{figure}
Diffusion models have emerged as the pre-eminent method for image generation, as evidenced by state-of-the-art approaches like Stable Diffusion \cite{rombach2022ldm}, Imagen \cite{saharia2022photorealistic}, eDiff-I \cite{balaji2023ediffi}, or Dall-E 2 \cite{ramesh2022dalle2}. They are versatile, succeeding in modalities such as video and audio \cite{blattmann2023videoldm,kong2021diffwave}. They boast scalability, training stability, and output diversity.

Diffusion model architectures employ diverse backbones, spanning CNN-based \cite{ho2020ddpm}, transformer-based \cite{peebles2023dit,bao2022uvit}, CNN-transformer-hybrid \cite{hoogeboom2023simple}, or even state-space models \cite{yan2023diffusionwithoutattention}. There is likewise variation in the approaches used to scale these models to support high-resolution image synthesis. Current approaches add complexity to training, necessitate additional models, or sacrifice quality.

Latent diffusion models \cite{rombach2022ldm} (LDMs) reign as the dominant method for achieving high-resolution image synthesis. In practice, they fail to represent fine detail \citep[see also \cref{fig:latent_quality_loss}]{dai2023emu}, impacting sample quality and limiting its utility in applications such as image editing. Other approaches to high-resolution synthesis include cascaded super-resolution \cite{saharia2022photorealistic}, multi-scale losses \cite{hoogeboom2023simple}, the incorporation of inputs and outputs at multiple resolutions \cite{gu2023matryoshka}, or the utilization of self-conditioning and the adaptation of fundamentally new architecture schemes \cite{jabri2023rin}.
\\

We seek to advance the state of pixel-space diffusion, offering a path to synthesis at high resolutions without resorting to LDMs. Eliminating the latent VAE frees us from quality limitations endemic to such VAEs (illustrated in \cref{fig:latent_quality_loss}), and bolsters downstream applications such as image editing (a process which LDMs encumber with poor reconstruction). We expound the merits of pixel-space DMs versus LDMs in \cref{sec:pixel_latent_comparison}.

{Our work tackles high-resolution synthesis via backbone improvements, which grant the efficiency needed to target pixel-space directly.}
We introduce a pure transformer architecture inspired by the hierarchical structure introduced in \cite{nawrot2022hourglass}, which we call the Hourglass Diffusion Transformer (\modelname{}). Our backbone is capable of high-quality image generation at megapixel scale in standard diffusion setups. This architecture, even at low spatial resolutions such as $128 \times 128$ is substantially more efficient than common diffusion transformer backbones such as DiT \cite{peebles2023dit} (see \cref{tab:main_ablation,fig:complexity_scaling}) while being competitive in generation quality. When scaling the model architecture to target resolutions according to our scheme, we obtain $\mathcal{O}(n)$ computational complexity scaling with the target number of image tokens $n$ in place of the $\mathcal{O}(n^2)$ scaling of normal diffusion transformer architectures, making this the first transformer-based diffusion backbone competitive in computational complexity with convolutional U-Nets for pixel-space high-resolution image synthesis.

\noindent Our main contributions are as follows:

\begin{itemize}
    \item We introduce the Hourglass Diffusion Transformer (\modelname{}), which achieves subquadratic scaling of compute with resolution. We show how our architecture choices help improve upon the quality of the baseline DiT \cite{peebles2023dit} in pixel-space image synthesis.
    \item We demonstrate high-quality pixel-space generation at $1024 \times 1024$ resolutions, setting a state-of-the-art FID for diffusion models on FFHQ-$1024^2$. We do so without training complications such as progressive growing or multiscale losses.
    \item We show HDiT's competence in a large-scaling training scenario through competitive evaluation on ImageNet-$256^2$. Quantitatively, it measures well against even latent transformer-based diffusion models despite undertaking the training at a higher effective resolution.
\end{itemize}

\begin{figure}[ht]
    \centering
    \adjustbox{max width=\columnwidth}{
    \input{img/vae_fidelity}
    }
    \caption{Motivation: Detail loss incurred through use of a standard VAE \cite{rombach2022ldm} on one of \cref{fig:teaser} samples. Notably, this VAE is employed by the baseline DiT \cite{peebles2023dit} architecture against which we compare.}
    \label{fig:latent_quality_loss}
\end{figure}

\begin{figure*}[htb]
    \centering
    \begin{adjustbox}{max width=\textwidth}
        \newcommand{\hditblock}[6]{
    \begin{scope}[rotate=-90]
        \node[standard node={white},minimum height=2.5mm,transform shape,minimum width=#5] at #2 (hditblocklayer1#1) {};

        \foreach \x in {2,...,#4} {
            \pgfmathtruncatemacro{\xlast}{\x - 1}
            \node[standard node={white},minimum height=2.5mm,above=2.5mm of hditblocklayer\xlast#1,minimum width=#5,transform shape] (hditblocklayer\x#1) {};
            \draw[standard arrow] (hditblocklayer\xlast#1) -- (hditblocklayer\x#1);
        }
        
        \begin{scope}[on background layer]
            \node[
                standard node={green},
                label={[align=center]above:{#3}},
                fit=(hditblocklayer1#1) (hditblocklayer1#1) (hditblocklayer#4#1) (hditblocklayer#4#1),
                minimum width=0,
                minimum height=0,
            ] (hditblock#1) {};
        \end{scope}
    \end{scope}
}
\newcommand{\tokenmergename}{2$\times$2 Pixel-unshuffle\\Merge + Proj.}
\newcommand{\tokensplitname}{Proj.+ 2$\times$2\\Pixel-shuffle Split}

\begin{tikzpicture}
    \begin{scope}[rotate=-90]
        \node[standard node={gray},minimum width=64mm,transform shape] (inputimage) {Image Input};
        \node[standard node={blue},minimum width=61mm,trapezium,trapezium stretches body,above=of inputimage,transform shape] (patching) {Patching ($p \times p$)\\+ Embedding};
        \draw[standard arrow] (inputimage) -- (patching);
    \end{scope}
    
    \hditblock{1}{($(patching) + (0,11mm)$)}{Neighborhood\\\modelname{} Blocks}{2}{47mm}{1}
    \draw[standard arrow] (patching) -- (hditblocklayer11);
    
    \begin{scope}[rotate=-90]
        \node[standard node={blue},minimum width=50mm,trapezium,trapezium stretches body,above=6mm of hditblock1.east,transform shape] (tokenmerge1) {\tokenmergename};
        \draw[standard arrow] (hditblocklayer21) -- (tokenmerge1);
    \end{scope}
    
    \hditblock{2}{($(tokenmerge1) + (0,11mm)$)}{Neighborhood\\\modelname{} Blocks}{2}{36mm}{2}
    \draw[standard arrow] (tokenmerge1) -- (hditblocklayer12);
    
    \begin{scope}[rotate=-90]
        \node[standard node={blue},minimum width=39mm,trapezium,trapezium stretches body,above=6mm of hditblock2.east,transform shape] (tokenmerge2) {\tokenmergename};
        \draw[standard arrow] (hditblocklayer22) -- (tokenmerge2);
    \end{scope}


    \hditblock{3}{($(tokenmerge2) + (0,11mm)$)}{Global\\\modelname{} Blocks}{5}{\tikzstyleminimumwidth}{3}
    \draw[standard arrow] (tokenmerge2) -- (hditblocklayer13);

    \begin{scope}[rotate=-90]
        \node[standard node={blue},minimum width=39mm,trapezium,trapezium left angle=120,trapezium right angle=120,trapezium stretches body,above=of hditblock3.east,transform shape] (tokensplit1) {\tokensplitname};
        \draw[standard arrow] (hditblocklayer53) -- (tokensplit1);
    \end{scope}
    \node[standard node circle={white}, right=of tokensplit1.north] (lerp1) {\small lerp};
    \draw[standard arrow] (tokensplit1) -- (lerp1);
    \draw[standard arrow] (hditblocklayer22.north) -- ++(4mm,0) |- ++(0,-22.5mm) -| (lerp1);
    
    \hditblock{4}{($(lerp1) + (0,10mm)$)}{Neighborhood\\\modelname{} Blocks}{2}{36mm}{2}
    \draw[standard arrow] (lerp1) -- (hditblocklayer14);
    
    \begin{scope}[rotate=-90]
        \node[standard node={blue},minimum width=50mm,trapezium,trapezium left angle=120,trapezium right angle=120,trapezium stretches body,above=of hditblock4.east,transform shape] (tokensplit2) {\tokensplitname};
        \draw[standard arrow] (hditblocklayer24) -- (tokensplit2);
    \end{scope}
    \node[standard node circle={white}, right=of tokensplit2.north] (lerp2) {\small lerp};
    \draw[standard arrow] (tokensplit2) -- (lerp2);
    \draw[standard arrow] (hditblocklayer21.north) -- ++(4mm,0) |- ++(0,-28mm) -| (lerp2);
    
    \hditblock{5}{($(lerp2) + (0,10mm)$)}{Neighborhood\\\modelname{} Blocks}{2}{47mm}{1}
    \draw[standard arrow] (lerp2) -- (hditblocklayer15);
    
    \begin{scope}[rotate=-90]
        \node[standard node={blue},above=of hditblock5.east,minimum width=50mm,transform shape] (outnorm) {RMSNorm};
        \draw[standard arrow] (hditblocklayer25) -- (outnorm);
        \node[standard node={blue},minimum width=61mm,trapezium,trapezium left angle=120,trapezium right angle=120,trapezium stretches body,above=of outnorm,transform shape] (tokensplit3) {Proj. + $p \times p$\\Pixel-shuffle};
        \draw[standard arrow] (outnorm) -- (tokensplit3);
        \node[standard node={gray},above=of tokensplit3,minimum width=64mm,transform shape] (outputimage) {Image Output};
        \draw[standard arrow] (tokensplit3) -- (outputimage);
    \end{scope}
\end{tikzpicture}
    \end{adjustbox}
    \caption{High-level overview of our \modelname{} architecture, specifically the version for ImageNet at input resolutions of $256^2$ at patch size $p = 4$, which has three levels. For any doubling in target resolution, another neighborhood attention block is added. ``lerp'' denotes a linear interpolation with learnable interpolation weight. All \modelname{} blocks have the noise level and the conditioning (embedded jointly using a mapping network) as additional inputs.}
    \vspace{-6mm}
    \label{fig:hdit_diagram}
\end{figure*}

\section{Related Work}\label{sec:related_work}
\subsection{Transformers}\label{sec:related_transformers}
The transformer architecture \cite{vaswani2017transformer} reigns as state-of-the-art in various domains \cite{openai2023gpt4,zong2022codetr,zhang2022pushing,yu2022coca,piergiovanni2023tubevit}. It has been scaled to tens of billions of parameters in the vision domain, \cite{dehghani2023scalingvit} and beyond that in natural language processing \cite{chowdhery2023palm,fedus2022switchtransformers}.
Transformers consider interactions between all elements in the sequence via the attention mechanism. Long-range interactions can be learned, but computational complexity scales quadratically with the length of input sequence.

\oursubsubsection{Transformer-based Diffusion Models}
Recent works have applied transformers to diffusion models. Diffusion priors \cite{ramesh2022dalle2} have provided low-dimensional embeddings on which to condition image synthesis, and latent diffusion \cite{rombach2022ldm} has achieved state-of-the-art performance generating images from compressed image latents \cite{peebles2023dit,bao2022uvit,zheng2023maskdit,gao2023mdt,bao2023unidiffusion,chen2023pixartalpha,chen2023gentron}. Transformer-based architectures \cite{hoogeboom2023simple,jing2023udittts} have been applied to U-Nets \cite{ronneberger2015unet}, either at the lowest levels \cite{ho2020ddpm}, or by altogether hybridizing the two architectures \cite{cao2022exploring}. The quadratic computational complexity of transformers' attention mechanism precludes high-resolution synthesis in pixel-space \cite{yang2022genvit}; latent representations are typically used to reduce the operating resolution.

Diffusion Transformers (DiT) \cite{peebles2023dit}, are amenable to masked training \cite{gao2023mdt,zheng2023maskdit}, which incentivizes models to better learn feature relationships. It is orthogonal and complementary to the architecture improvements pursued in this work.

\oursubsubsection{Hourglass Transformers}
The Hourglass architecture \cite{nawrot2022hourglass} is a hierarchical implementation of transformers, shown to be more efficient at language modeling than standard Transformer models, in training and in inference. Sequences are shortened as they descend the encoder levels of the transformer, culminating in the shortest representation in the middle, then re-expanded as they ascend the decoder levels. Skip connections reintroduce higher-resolution information near the expansion steps. Hourglasses resemble U-Nets \cite{ronneberger2015unet} without convolutional layers. Hierarchical structures \cite{wang2021uformer} have excelled at image restoration, a task similar to the denoising objective pursued in diffusion.

\subsection{High-Resolution Image Synthesis with Diffusion Models}
High-resolution image synthesis in diffusion models has been extensively studied, yet it remains a challenge to current single-stage models. Popular approaches separate the generation process into multiple steps. Cascaded super-resolution \cite{ho2021cascaded} targets initially a low-resolution image, scaling it via a series of super-resolution models. Latent diffusion targets a spatially downsampled ``latent'' representation, which can be decoded into a higher-resolution pixel image via a convolutional model \cite{rombach2022ldm} or another diffusion model \cite{betker2023dalle3}. The latent representation can itself also be super-resoluted \cite{fischer2023boosting}. Latent diffusion is the strategy chosen by most transformer-based diffusion models (see \cref{sec:related_transformers}). Recent works explore high-resolution image synthesis in pixel space, in an effort to simplify the overall architecture. Fundamentally new backbone architectures \cite{jabri2023rin} have been proposed. Spatial dimensions have been reduced via discrete wavelet transforms \cite{hoogeboom2023simple}. The diffusion training process has not stood still, with proposals such as self-conditioning across sampling steps \cite{jabri2023rin}, multiresolution training \cite{gu2023matryoshka}, and multiresolution losses \cite{hoogeboom2023simple} offering a path to higher resolutions. The necessity of such substantial modifications of the diffusion process is proving difficult to overcome, with simpler approaches \cite{song2021scorebased} -- single-stage and lacking the aforementioned training adaptations -- struggling to produce samples that fully utilize the available resolution and are globally coherent.

\section{Preliminaries}\label{sec:preliminaries}
\subsection{Diffusion Models}
Diffusion models generate data by learning to reverse a diffusion process. This diffusion process is most commonly defined to be a Gaussian noising process. Given a data distribution $p_\text{data}(\mathbf{x})$, we define a \textit{forward} noising process with the family of distributions $p(\mathbf{x}_{\sigma_t};{\sigma_t})$ that is obtained by adding i.i.d. Gaussian noise of standard deviation ${\sigma_t}$ which is provided by a predefined monotonically increasing noise level schedule.  Therefore, $\mathbf{x}_{\sigma_t} = \mathbf{x}_0 + \sigma_t \epsilon$ where $\mathbf{\epsilon} \sim \mathcal{N}\left(\mathbf{0}, \mathbf{I}\right)$. A denoising neural network $D_\theta\left(\mathbf{x}_{\sigma_t}, {\sigma_t}\right)$ is trained to predict $\mathbf{x}_0$ given $\mathbf{x}_{\sigma_t}$. Sampling is done by starting at $\mathbf{x}_T \sim \mathcal{N}\left(\mathbf{0}, \sigma_\text{max}^2\mathbf{I}\right)$ and sequentially denoising at each of the noise levels before resulting in the sample $\mathbf{x}$. The denoiser neural network is trained with a mean-squared error loss:
\begin{equation}
\mathbb{E}_{\mathbf{x} \sim p_\text{data}(\mathbf{x})} \mathbb{E}_{\epsilon,{\sigma_t} \sim p(\epsilon,{\sigma_t})} \left[\lambda_{\sigma_t} \|D_\theta\left(\mathbf{x}_{\sigma_t}, {\sigma_t}\right) - \mathbf{x} \|^2_2\right],
\end{equation}
where $\lambda_{\sigma_t}$ is a weighting function.

Recent works proposed various improvements to this basic formulation. Two notable approaches, which are also adapted by our model, are preconditioning to obtain more suitable prediction targets for the model \cite{karras2022elucidating} and adapting the loss weighting to a clamped signal-to-noise ratio (SNR) $\lambda_{\sigma_t} = \min\{\frac{1}{\sigma_t}, \gamma\}$ to improve model convergence \cite{hang2023minsnr}. Another improvement has been the adaption of noise schedules for high resolutions. It was previously observed \cite{hoogeboom2023simple} that the commonly used noise schedules that were originally designed for low resolutions (32x32 or 64x64) fail to add enough noise at high resolutions. Therefore, the noise schedules can be shifted and interpolated from a reference low-resolution noise schedule in order to add appropriate noise at higher resolutions.

\section{Hourglass Diffusion Transformers}\label{sec:hourglass}
Diffusion Transformers \cite{peebles2023dit} and other similar works (see \cref{sec:related_transformers}) have demonstrated impressive performance as denoising diffusion autoencoders in latent diffusion \cite{rombach2022ldm} setups, surpassing prior works in terms of generative quality \cite{gao2023mdt,zheng2023maskdit}. However, their scalability to high resolutions is limited by the fact that the computational complexity increases quadratically ($\mathcal{O}(n^2)$ for images of shape $h \times w \times \text{channels}$, with $n = w \cdot h$). This makes them prohibitively expensive to train and run on high-resolution inputs, effectively limiting transformers to spatially compressed latents at sufficiently small dimensions, unless very large patch sizes are used \cite{cao2022exploring}, which have been found to be detrimental to the quality of generated samples \cite{peebles2023dit}.

We propose a new, improved hierarchical architecture based on Diffusion Transformers \cite{peebles2023dit}, and Hourglass Transformers \cite{nawrot2022hourglass} -- Hourglass Diffusion Transformers (\modelname{}) -- that enables high-quality pixel-space image generation and can be efficiently adapted to higher resolutions with a computational complexity scaling of $\mathcal{O}(n)$ instead of $\mathcal{O}(n^2)$.
This means that even scaling up these models to direct pixel-space generation at megapixel resolutions becomes viable, which we demonstrate for models at resolutions of up to $1024 \times 1024$ in \cref{sec:experiments}.

\subsection{Leveraging the Hierarchical Nature of Images}\label{sec:hourglass_structure_diffusion}
Natural images exhibit hierarchies \cite{Saremi2013}. This makes mapping the image generation process into a hierarchical model an intuitive choice, which has previously been successfully applied in the U-Net architecture \cite{ronneberger2015unet} commonly used in diffusion models but is not commonly used by diffusion transformers \cite{peebles2023dit,bao2022uvit}. To leverage this hierarchical nature of images for our transformer backbone, we apply the hourglass structure \cite{nawrot2022hourglass}, which has been shown to be effective for a range of different modalities, including images, for the high-level structure of our transformer backbone. 
Based on the model's primary resolution, we choose the number of levels in the hierarchy, such that the innermost level has $16 \times 16$ tokens. We use a larger hidden dimension for lower-resolution levels, which have to process both low-resolution information and information relevant for following higher-resolution levels. For every level on the encoder side, we spatially merge $2 \times 2$ tokens into one using Pixel-UnShuffle \cite{shi2016pixelshuffle} and do the inverse on the decoder side.

\oursubsubsection{Skip Merging Mechanism}
One important consideration in hierarchical architectures is the merging mechanisms of skip connections, as it can influence the final performance significantly \cite{bao2022uvit}. While the previous non-hierarchical U-ViT \cite{bao2022uvit} uses a concatenation-based skip implementation, similar to the standard U-Net \cite{ronneberger2015unet}, and found this to be significantly better than other options, we find additive skips to perform better for this hierarchical architecture. As the usefulness of the information provided by the skips can differ significantly, especially in very deep hierarchies, we additionally enable the model to learn the relative importance of the skip and the upsampled branch by learning a linear interpolation (lerp) coefficient $f$ between the two for each skip:
\begin{equation}
    \mathbf{x}_\mathrm{merged}^\text{(l. lerp)} = f \cdot \mathbf{x}_\mathrm{skip} + (1 - f) \cdot \mathbf{x}_\mathrm{upsampled}.
\end{equation}

\subsection{Hourglass Diffusion Transformer Block Design}\label{sec:block_design}
\begin{figure}[ht]
    \centering
    \begin{subfigure}[h]{.5\columnwidth}
        \centering
        \vspace{8.9mm}
        \scalebox{.65}{%
            \centering
            \begin{tikzpicture}
    \node[standard node={gray}] (inputtokens) {Input Tokens};
    \node[standard node={gray}, right=of inputtokens] (conditioninginput) {Conditioning};
    \draw[standard arrow] ($(inputtokens.south) + (0,-.4cm)$) -- (inputtokens);
    \draw[standard arrow] ($(conditioninginput.south) + (0,-.4cm)$) -- (conditioninginput);

    \node[standard node={blue}, above=7mm of inputtokens] (selfattentionnorm) {AdaRMSNorm};
    \draw[standard arrow] (inputtokens) -- (selfattentionnorm);
    \node[standard node={yellow}, above=of selfattentionnorm] (mhsa) {Multi-Head RoPE\\Cosine Similarity\\Self-Attention};
    \draw[standard arrow] (selfattentionnorm) -- (mhsa);
    \node[standard node circle={white}, above=of mhsa] (selfattentionskip) {+};
    \draw[standard arrow] (mhsa) -- (selfattentionskip);
    \draw[standard arrow] (inputtokens.north) -- ++(0,0.15cm) -| ++(-1.55cm,0) |- ($(selfattentionskip.west) + (-.1cm,0)$);

    \node[standard node={blue}, above=of selfattentionskip] (ffnnorm) {AdaRMSNorm};
    \draw[standard arrow] (selfattentionskip) -- (ffnnorm);
    \node[standard node={green}, above=of ffnnorm] (ffnpointwise) {\modelname{} Pointwise\\Feedforward};
    \draw[standard arrow] (ffnnorm) -- (ffnpointwise);
    \node[standard node circle={white}, above=of ffnpointwise] (ffnskip) {+};
    \draw[standard arrow] (ffnpointwise) -- (ffnskip);
    \draw[standard arrow] (selfattentionskip.north) -- ++(0,0.15cm) -| ++(-1.55cm,0) |- ($(ffnskip.west) + (-.1cm,0)$) node [midway] (skiparrowmidway) {};
    \draw[standard arrow] (ffnskip.north) -- ++(0,0.4cm);
    
    \node[standard node={blue}, above=of conditioninginput] (conditioningmlp) {MLP};
    \draw[standard arrow] (conditioninginput) -- (conditioningmlp);
    \draw[standard arrow] (conditioningmlp.north) |- node[above,pos=0.8] {$\gamma_1$} (selfattentionnorm.east);
    \draw[standard arrow] (conditioningmlp.north) |- node[above,pos=0.8] {$\gamma_2$} (ffnnorm.east);

    \begin{scope}[on background layer]
        \node[
            standard node={white},
            fit=(ffnskip) (inputtokens) (skiparrowmidway) (conditioninginput)
        ] {};
    \end{scope}
\end{tikzpicture}
        }
        \caption{\modelname{} Block Architecture.}
    \end{subfigure}%
    \begin{subfigure}[h]{.5\columnwidth}
        \centering
        \scalebox{.65}{%
            \centering
            \begin{tikzpicture}
    \node[standard node={gray}] (inputtokens) {Input Tokens};
    \node[standard node={gray}, right=of inputtokens] (conditioninginput) {Conditioning};
    \draw[standard arrow] ($(inputtokens.south) + (0,-.4cm)$) -- (inputtokens);
    \draw[standard arrow] ($(conditioninginput.south) + (0,-.4cm)$) -- (conditioninginput);

    \node[standard node={blue}, above=7mm of inputtokens] (selfattentionnorm) {AdaLN};
    \draw[standard arrow] (inputtokens) -- (selfattentionnorm);
    \node[standard node={yellow}, above=of selfattentionnorm] (mhsa) {Multi-Head\\Self-Attention};
    \draw[standard arrow] (selfattentionnorm) -- (mhsa);
    \node[standard node={blue}, above=of mhsa] (selfattentionscale) {Scale};
    \draw[standard arrow] (mhsa) -- (selfattentionscale);
    \node[standard node circle={white}, above=of selfattentionscale] (selfattentionskip) {+};
    \draw[standard arrow] (selfattentionscale) -- (selfattentionskip);
    \draw[standard arrow] (inputtokens.north) -- ++(0,0.15cm) -| ++(-1.55cm,0) |- ($(selfattentionskip.west) + (-.1cm,0)$) node [midway] (skiparrowmidway) {};

    \node[standard node={blue}, above=of selfattentionskip] (ffnnorm) {AdaLN};
    \draw[standard arrow] (selfattentionskip) -- (ffnnorm);
    \node[standard node={green}, above=of ffnnorm] (ffnpointwise) {DiT Pointwise\\Feedforward};
    \draw[standard arrow] (ffnnorm) -- (ffnpointwise);
    \node[standard node={blue}, above=of ffnpointwise] (ffnscale) {Scale};
    \draw[standard arrow] (ffnpointwise) -- (ffnscale);
    \node[standard node circle={white}, above=of ffnscale] (ffnskip) {+};
    \draw[standard arrow] (ffnscale) -- (ffnskip);
    \draw[standard arrow] (selfattentionskip.north) -- ++(0,0.15cm) -| ++(-1.55cm,0) |- ($(ffnskip.west) + (-.1cm,0)$);
    \draw[standard arrow] (ffnskip.north) -- ++(0,0.4cm);
    
    \node[standard node={blue}, above=of conditioninginput] (conditioningmlp) {MLP};
    \draw[standard arrow] (conditioninginput) -- (conditioningmlp);
    \draw[standard arrow] (conditioningmlp.north) |- node[above,pos=0.8] {$\gamma_1,\beta_1$} (selfattentionnorm.east);
    \draw[standard arrow] (conditioningmlp.north) |- node[above,pos=0.86] {$\alpha_1$} (selfattentionscale.east);
    \draw[standard arrow] (conditioningmlp.north) |- node[above,pos=0.8] {$\gamma_2,\beta_2$} (ffnnorm.east);
    \draw[standard arrow] (conditioningmlp.north) |- node[above,pos=0.86] {$\alpha_2$} (ffnscale.east);

    \begin{scope}[on background layer]
        \node[
            standard node={white},
            fit=(ffnskip) (inputtokens) (skiparrowmidway) (conditioninginput)
        ] {};
    \end{scope}
\end{tikzpicture}
        }
        \caption{DiT Block Architecture.}
    \end{subfigure}
    \caption{A comparison of our transformer block architecture and that used by DiT \cite{peebles2023dit}.}
    {\vspace{-1.5mm}}
    \label{fig:hdit_block_diagram}
\end{figure}

\begin{figure}[htb]
    \centering
    \begin{subfigure}[h]{.5\columnwidth}
        \centering
        \scalebox{.65}{%
            \begin{tikzpicture}
    \node[standard node={gray}] (inputtokens) {Input};
    \draw[standard arrow] ($(inputtokens.south) + (0,-.4cm)$) -- (inputtokens);

    \node[above=of inputtokens,minimum width=0cm,minimum height=9mm] (linearcenter) {};
    \node[standard node={green},minimum width=1cm,left=of linearcenter.east] (linear1) {Linear};
    \node[standard node={green},minimum width=1cm,right=of linearcenter.west] (linear2) {Linear};
    \draw[standard arrow] (inputtokens.north) -- ++(0,0.15cm) -| (linear2);
    \draw[standard arrow] (inputtokens.north) -- ++(0,0.15cm) -| (linear1);
    
    \node[standard node={blue},minimum width=1cm,above=of linear1] (gelu) {GELU};
    \draw[standard arrow] (linear1) -- (gelu);
    \node[above=of linearcenter,minimum width=0cm,minimum height=3mm] (linearcenter2) {};
    
    \node[standard node circle={white}, above=of linearcenter2] (prod) {$\odot$};
    \draw[standard arrow] (gelu) |- (prod);
    \draw[standard arrow] (linear2) |- (prod);
    \node[standard node={blue},dashed, above=of prod] (dropout) {Dropout};
    \draw[standard arrow] (prod) -- (dropout);
    \node[standard node={green}, above=of dropout] (downproj) {Linear};
    \draw[standard arrow] (dropout) -- (downproj);
    \node[standard node circle={white}, above=of downproj] (ffnskip) {+};
    \draw[standard arrow] (downproj) -- (ffnskip);
    \draw[standard arrow] (inputtokens.north) -- ++(0,0.15cm) -| ++(-1.9cm,0) |- ($(ffnskip.west) + (-.1cm,0)$) node [midway] (skiparrowmidway) {};
    
    \draw[standard arrow] (ffnskip.north) -- ++(0,0.4cm);

    \node[right=of linear2] (rightspacing) {};
    \begin{scope}[on background layer] 
        \node[
            standard node={white},
            fit=(ffnskip) (skiparrowmidway) (rightspacing) (inputtokens)
        ] {};
    \end{scope}
    
    \begin{scope}[on background layer] 
        \node[
            standard node={white},
            label={[rotate=90,xshift=-1.85cm,yshift=-2.5mm]right:{GEGLU \cite{shazeer2020glu}}},
            fit=(prod) (linear1) (linear1) (linear2)
        ] {};
    \end{scope}
\end{tikzpicture}
        }
        \caption{\modelname{} FFN Block.}
    \end{subfigure}%
    ~
    \begin{subfigure}[h]{.3\columnwidth}
        \centering
        \vspace{6.5mm}
        \scalebox{.65}{%
            \begin{tikzpicture}
    \node[standard node={gray}] (inputtokens) {Input};
    \draw[standard arrow] ($(inputtokens.south) + (0,-.4cm)$) -- (inputtokens);

    \node[standard node={green}, above=of inputtokens] (linear) {Linear};
    \draw[standard arrow] (inputtokens) -- (linear);

    \node[standard node={blue}, above=of linear] (act1) {GELU};
    \draw[standard arrow] (linear) -- (act1);

    \node[standard node={green}, above=of act1] (downproj) {Linear};
    \draw[standard arrow] (act1) -- (downproj);
    \node[standard node circle={white}, above=of downproj] (ffnskip) {+};
    \draw[standard arrow] (downproj) -- (ffnskip);
    \draw[standard arrow] (inputtokens.north) -- ++(0,0.15cm) -| ++(-1.55cm,0) |- ($(ffnskip.west) + (-.1cm,0)$) node [midway] (skiparrowmidway) {};
    
    \draw[standard arrow] (ffnskip.north) -- ++(0,0.4cm);

    \begin{scope}[on background layer]
        \node[
            standard node={white},
            fit=(ffnskip) (skiparrowmidway) (inputtokens) (inputtokens)
        ] {};
    \end{scope}
\end{tikzpicture}
        }
        \vspace{6mm}
        \caption{DiT FFN Block.}
    \end{subfigure}
    \caption{A comparison of our pointwise feedforward block architecture and that used by DiT \cite{peebles2023dit}.}
    {\vspace{-6mm}}
    \label{fig:ffn_diagram}
\end{figure}

Our basic transformer block design (shown in comparison with that of DiT in \cref{fig:hdit_block_diagram}) is generally inspired by the blocks used by LLaMA \cite{touvron2023llama}, a transformer architecture that has recently been shown to be very capable of high-quality generation of language.
To enable conditioning, we make the output scale used by RMSNorm operations adaptive, predicted by a mapping network conditioned on the class and diffusion time step.
Unlike DiT, we do not employ an (adaptive) output gate, but initialize the output projections of both self-attention and FFN blocks to zeros.
To make positional information accessible to the transformer model, common diffusion transformer architectures like DiT and U-ViT use a learnable additive positional encoding. \cite{peebles2023dit,bao2022uvit} As it is known to improve models' generalization and their capability of extrapolating to new sequence lengths, we replace this with an adaptation of rotary positional embeddings (RoPE) \cite{su2022roformer} for 2D image data: we follow an approach similar to \cite{ho2019axial} and split the encoding to operate on each axis separately, applying RoPE for each spatial axis to distinct parts of query and key respectively. We also found that applying this encoding scheme to only half of the query and key vectors and not modifying the rest to be beneficial for performance. Overall, we find empirically that replacing the normal additive positional embedding with our adapted RoPE improves convergence and helps remove patch artifacts.
Additionally to applying RoPE, we use a cosine similarity-based attention mechanism that has previously been used in \cite{liu2022swin} (see \cref{sec:cosine_sim_attn} for details). We note that a similar approach has been proven at the multi-billion parameter scale for vision transformers \cite{dehghani2023scalingvit}.

\noindent For the feedforward block (see \cref{fig:ffn_diagram} for a comparison with DiT), instead of having an output gate like DiT, we use GEGLU \cite{shazeer2020glu}, where the modulation signal comes from the data itself instead of the conditioning and is applied on the first instead of the second layer of the FFN.

\subsection{Efficient Scaling to High Resolutions}
The hourglass structure enables us to process an image at a variety of resolutions. We use global self-attention at low resolutions to achieve coherence, and local self-attention \cite{liu2021swin,liu2022swin,hassani2023neighborhood} at all higher resolutions to enhance detail. This limits the need for quadratic-complexity global attention to a manageable amount, and enjoys linear-complexity scaling for any further increase in resolution. Asymptotically, the complexity is $\mathcal{O}(n)$ (see \cref{sec:computational_complexity}) w.r.t pixel count $n$.

A typical choice for localized self-attention would be Shifted Window attention \cite{liu2021swin,liu2022swin} as used by previous diffusion models \cite{cao2022exploring,li2022swinv2imagen}. We find, however, that Neighborhood attention \cite{hassani2023neighborhood} performs significantly better in practice.

The maximum resolution at which to apply global self-attention\footnote{For our FFHQ-$1024^2$ experiment, we apply two levels of global attention -- one at $16^2$ and one at $32^2$. Whereas for ImageNet-$128^2$ and $256^2$, we found like prior works \cite{ho2020ddpm,hoogeboom2023simple,nichol2021iddpm} that a single level of $16^2$ global attention suffices.} is a choice determined by dataset (the size at which small features requiring long-distance coherence become large enough for attention to reason about) and by task (the smallest feature whose long-distance relationships need to be preserved in order to be acceptable). At particularly low resolutions (e.g. $256^2$), some datasets permit coherent generation with fewer levels of global attention.

\section{Experiments}\label{sec:experiments}
\begin{minipage}{\columnwidth}

We evaluate the proposed \modelname{} architecture on conditional and unconditional image generation, ablating over architectural choices (\cref{sec:ablation_study}), and evaluating both megapixel pixel-space image generation (\cref{sec:high_res_image_synthesis}) and large-scale pixel-space image generation (\cref{sec:large_scale_imagenet}).

\subsection{Experimental Setup}
\vspace{0.14in}
\end{minipage}
\oursubsubsection{Training}
Unless mentioned otherwise, we train class-conditional models on ImageNet \cite{jia2009imagenet} at a resolution of $128 \times 128$ directly on RGB pixels without any kind of latent representation.
We adapt our general training setup from \cite{karras2022elucidating}, including their preconditioner, and use a continuous-time diffusion formulation.
We train all models with AdamW \cite{loshchilov2018adamw} using a constant learning rate of $5 \times 10^{-4}$ and a weight decay of $\lambda = 0.01$. We generally train at a batch size of $256$ for 400k steps (following \cite{peebles2023dit}) with stratified diffusion timestep sampling and do not use Dropout unless noted otherwise.
For small-scale ImageNet trainings at $128 \times 128$, we do not apply any augmentation. For runs on small datasets, we apply a non-leaking augmentation scheme akin to \cite{karras2020styleganada}.
Following common diffusion model training practice and \cite{peebles2023dit}, we also compute the exponential moving average (EMA) of the model weights with a decay of $0.9999$. We use this EMA version of the model for all evaluations and generated samples, and perform our sampling using 50 steps of DPM++(3M) \cite{lu2023dpmsolver} SDE sampling. For further details, see \cref{tab:training_details}.

\oursubsubsection{Evaluation}
Following common practice for generative image models, we report the Fr\'echet Inception Distance (FID) \cite{heusel2017fid} computed on 50k samples. To compute FID, we use the commonly used implementation
from \cite{dhariwal2021diffusion}.
We also report both the absolute and asymptotic computational complexity for our main ablation study, also including FLOPs for higher-resolution versions of the architecture.

\subsection{Effect of the Architecture}\label{sec:ablation_study}
\newcommand{\ablationidbasic}{\textbf{A}}
\newcommand{\ablationidswi}{\textbf{B1}}
\newcommand{\ablationidnatten}{\textbf{B2}}
\newcommand{\ablationidgeglu}{\textbf{C}}
\newcommand{\ablationidrope}{\textbf{D}}
\newcommand{\ablationidmss}{\textbf{E}}
\newcommand{\ablationiddit}{\textbf{R1}}
\newcommand{\ablationidditoursetup}{\textbf{R2}}
\newcommand{\ablationidditourblocks}{\textbf{R3}}
\newcommand{\ablationidditmss}{\textbf{R4}}
To evaluate the effect of our architectural choices, we perform an ablation study where we start with a basic implementation of the hourglass architecture for diffusion and iteratively add the changes that enable our final architecture to efficiently perform high-quality megapixel image synthesis. We denote the ablation steps as \ablationidbasic, \ablationidswi, ..., \ablationidmss, and show their feature composition and experimental results in \cref*{tab:main_ablation}. We also provide a set of baselines \ablationiddit, ..., \ablationidditmss, where we trained DiT \cite{peebles2023dit} models in various settings to enable a fair comparison. Additional experimental steps are shown in \cref{sec:additional_ablations}.

\begin{table*}[htb]
    \centering
    \caption{
        Ablation of our architectural choices, starting from a stripped-down implementation of our hourglass diffusion transformer that is similar to DiT-B/4 \cite{peebles2023dit}. We also ablate over our additional choice of using soft-min-snr loss weighting, which we use to train our full models but do not consider part of our architecture. We present results for various DiT-B/4-based models as baselines. We also report computational cost per forward pass at multiple resolutions, including standard resolution-dependent model adaptations (relative to \ablationiddit{} in {\color{gray}gray}). See \cref{tab:additional_ablations} for an additional results.
    }
    \newcommand{\ablationtablestepref}[7]{#1 & #2 & #3 & #4 & #5 & #6 {\color{white}($-00\%$)} & #7 {\color{white}($-00\%$)}}
    \newcommand{\ablationtablestep}[9]{#1 & #2 & #3 & #4 & #5 & #6 {\color{gray}(#7)} & #8 {\color{gray}(#9)}}
    \newcommand{\boldon}{$\mathbf{\mathbfcal{O}\boldsymbol{(}\boldsymbol{n}\boldsymbol{)}}$}
    \begin{adjustbox}{max width=\textwidth}
        \begin{tabular}{l@{\hskip 1.5mm}|l|ccc||rr}\toprule
            \multicolumn{2}{l|}{\textbf{Configuration}} & \shortstack{\textbf{FID}\lowerisbetter} & \textbf{GFLOP@}$128^2$\lowerisbetter & \textbf{Complexity\lowerisbetter} & \textbf{GFLOP@}$256^2$ & \textbf{GFLOP@}$512^2$ \\
            \midrule
            \multicolumn{7}{l}{\textbf{Baselines} (\ablationiddit{} uses 250 DDPM sampling steps with learned $\sigma(t)$ as in the original publication instead of 50-step DPM++ sampling)}\\
            \ablationtablestepref{\ablationiddit{} }{DiT-B/4 \cite{peebles2023dit}}{42.03}{106}{$\mathcal{O}(n^2)$}{657}{6,341}\\
            \ablationtablestepref{\ablationidditourblocks{} }{\ablationiddit{} + our basic blocks \& mapping net \& trainer}{42.49}{106}{$\mathcal{O}(n^2)$}{657}{6,341}\\
            \midrule
            \ablationtablestepref{\ablationidditmss{} }{\ablationidditourblocks{} + Soft-Min-SNR}{\underline{30.71}}{106}{$\mathcal{O}(n^2)$}{657}{6,341}\\
            \midrule
            \midrule
            \multicolumn{7}{l}{\textbf{Ablation Steps}}\\
            \ablationtablestep{\ablationidbasic{} }{Global Attention Diffusion Hourglass (\cref{sec:hourglass_structure_diffusion})}{50.76}{{\color{white}0}32}{$\mathcal{O}(n^2)$}{114}{$-83\%$}{1,060}{$-83\%$}\\
            \ablationtablestep{\ablationidswi{} }{\ablationidbasic{} + Swin Attn. \cite{liu2021swin}}{55.93}{{\color{white}0}\textbf{29}}{\boldon}{{\color{white}0}60}{$-91\%$}{{\color{white}0,}185}{$-97\%$}\\
            \ablationtablestep{\ablationidnatten{} }{\ablationidbasic{} + Neighborhood Attn. \cite{hassani2023neighborhood}}{51.07}{{\color{white}0}\textbf{29}}{\boldon}{{\color{white}0}60}{$-91\%$}{{\color{white}0,}184}{$-97\%$}\\
            \ablationtablestep{\ablationidgeglu{} }{\ablationidnatten{} + GeGLU \cite{shazeer2020glu}}{44.36}{{\color{white}0}\underline{31}}{\boldon}{{\color{white}0}65}{$-90\%$}{{\color{white}0,}198}{$-96\%$}\\
            \ablationtablestep{\ablationidrope{} }{\ablationidgeglu{} + Axial RoPE (\cref{sec:block_design})}{41.41}{{\color{white}0}\underline{31}}{\boldon}{{\color{white}0}65}{$-90\%$}{{\color{white}0,}198}{$-96\%$}\\
            \midrule
            \ablationtablestep{\ablationidmss{} }{\ablationidrope{} + Soft-Min-SNR (\cref{sec:min_soft_snr})}{\textbf{27.74}}{{\color{white}0}\underline{31}}{\boldon}{{\color{white}0}65}{$-90\%$}{{\color{white}0,}198}{$-96\%$}\\
            \bottomrule
        \end{tabular}
    \end{adjustbox}
    \vspace{-3mm}
\label{tab:main_ablation}
\end{table*}

We generally use DiT-B-scale models for this comparison (approx. 130M parameters for DiT, approx 105M to 120M for \modelname{} depending on the ablation step), due to their relatively low training cost, and train them on pixel-space ImageNet \cite{jia2009imagenet} at a resolution of $128^2$ and patch size of 4.
The computational cost for the same architecture at resolutions of $256 \times 256$ and $512 \times 512$ is also reported. In the case of our models, every doubling in resolution involves adding one local attention block (except for ablation step \ablationidbasic{},
where it is global) as per \cref{sec:hourglass_structure_diffusion}.

\oursubsubsection{Baselines}
We train multiple versions of DiT in different setups to provide fair comparisons with it as baselines in \cref{tab:main_ablation}. \ablationiddit{} directly uses the official DiT implementation \cite{peebles2023dit} but omits the VAE latent computation step and adjusts the scaling to fit the data. No other changes were made, as DiT can be directly applied to pixel space \cite{peebles2023dit}.
We also train a baseline \ablationidditourblocks{} that uses the DiT-B hyperparameters and structure but applies them to our block architecture and training setup as used in \ablationidbasic{}. This matches the performance of the original DiT trained with the original codebase. On top of this setup, we also add soft-min-snr loss weighting to \ablationidditmss{} (as in ablation step \ablationidmss{}) to enable a fair comparison with our final model.

\oursubsubsection{Base Hourglass Structure}
Configuration \ablationidbasic{} is a simple hourglass structure with lower-resolution levels and our linear skip interpolations, and the basic implementation of our blocks with RMSNorm, but without GEGLU, and with full global self-attention at every level. A simple additive positional encoding is used here.
Even this simple architecture, without any of our additional changes, is already substantially cheaper (30\% of the FLOPs per forward pass, less for higher resolutions) than similarly-sized DiT \cite{peebles2023dit} models operating in pixel space due to the hourglass structure. For higher resolutions than $128^2$, this makes it viable to train pixel-space transformer-based models at all. This comes at the cost of increased FID compared to the DiT baselines at this step in the ablation.

\oursubsubsection{Local Attention Mechanism}
Next, we add local attention to all levels except for the lowest-resolution one. We evaluate two options -- Shifted-Window (Swin) \cite{liu2021swin,liu2022swin} attention (\ablationidswi{}, a common choice in vision transformers and previously also used in diffusion models \cite{cao2022exploring,li2022swinv2imagen}) and Neighborhood \cite{hassani2023neighborhood} attention (\ablationidnatten{}). Both result in a small reduction in FLOPs even at the low-resolution scale of $128 \times 128$ but, most importantly, reduce the computational complexity w.r.t. the base resolution from $\mathcal{O}(n^2)$ to $\mathcal{O}(n)$, enabling practical scaling to significantly higher resolutions.
Both variants suffer from increased FID due to this reduced expressiveness of local attention. Still, this change is significantly less pronounced for Neighborhood attention, making it a clearly superior choice in this case compared to the common choice of Swin attention.

\oursubsubsection{Feedforward Activation}
As the third step, we ablate over using GEGLU \cite{shazeer2020glu}, where the data itself affects the modulation of the outputs of the feedforward block, compared to the standard GeLU for the feedforward network. Similar to previous work \cite{touvron2023llama}, to account for the effective change of the hidden size due to the GEGLU operation, we decrease the hidden dimension from $4 \cdot d_\mathrm{model}$ to $3 \cdot d_\mathrm{model}$. We find that this change significantly improves FID at the cost of a slight increase in computational cost, as the width of the linear projections in the feedforward block has to be increased to account for the halving in output width.

\oursubsubsection{Positional Encoding}
Next, we replace the standard additive positional embedding with our 2D axial adaptation of RoPE \cite{su2022roformer} in \ablationidrope{} (see \cref{sec:axial_rope_details} for details), completing our Hourglass DiT backbone architecture. This further improves FID. As an additional benefit, RoPE should enable significantly better extrapolation to other resolutions than additive positional embeddings, but this ablation study does not test for that. Qualitatively, we find that this also helps reduce patching artifacts in the generated images.

\oursubsubsection{Loss Weighting}
Finally, we also ablate over replacing the standard $\frac{1}{\sigma^2}$ loss weighting \cite{ho2020ddpm,song2021scorebased} with our adapted Min-SNR \cite{hang2023minsnr} loss weighting method that we call Soft-Min-SNR (see \cref{sec:min_soft_snr}), which reduces the loss weight compared to SNR weighting for low noise levels. This substantially improves FID further, demonstrating the effectiveness of \modelname{} when coupled with an appropriate training setup for pixel-space diffusion.

\oursubsubsection{Skip Implementation}
Additionally to the main ablation study, we also ablate over different skip implementations based on ablation step \ablationidmss{}. We compare our learnable \ul{l}inear int\ul{erp}olation (lerp), which we empirically found to be especially helpful when training deep hierarchies, with both a standard additive skip, where the upsampled and skip data are directly added, and a concatenation version, where the data is first concatenated and then projected to the original channel count using a pointwise convolution. The results of this ablation are shown in \cref{tab:skip_ablation}.
We find that, even for shallow hierarchies as used for ImageNet-$128^2$ generation in our ablations, the learnable linear interpolation outperforms the addition slightly, with both the learnable lerp and addition substantially outperforming the commonly used concatenation.

\begin{table}[tb]
    \centering
    \caption{Skip Information Merging Mechanism Ablation}
    \begin{adjustbox}{max width=\columnwidth}
        \begin{tabular}{lc}\toprule
            \textbf{Skip Implementation} & \textbf{FID}\lowerisbetter \\
            \midrule
            Concatenation (U-Net \cite{ronneberger2015unet}) & 33.75 \\
            Addition (Original Hourglass \cite{nawrot2022hourglass}) & \underline{28.37} \\
            Learnable Linear Interpolation (\textbf{Ours}) & \textbf{27.74} \\
            \bottomrule
        \end{tabular}
    \end{adjustbox}
    \label{tab:skip_ablation}
\end{table}

\subsection{High-Resolution Pixel-Space Image Synthesis}\label{sec:high_res_image_synthesis}
In this section, we train our model for high-resolution pixel-space image synthesis. Following previous works, we train on FFHQ-$1024^2$ \cite{karras2021stylegan}, the standard benchmark dataset for image generation at such high resolutions.

\begin{figure}[tb]
    \centering
    \includegraphics[width=\columnwidth]{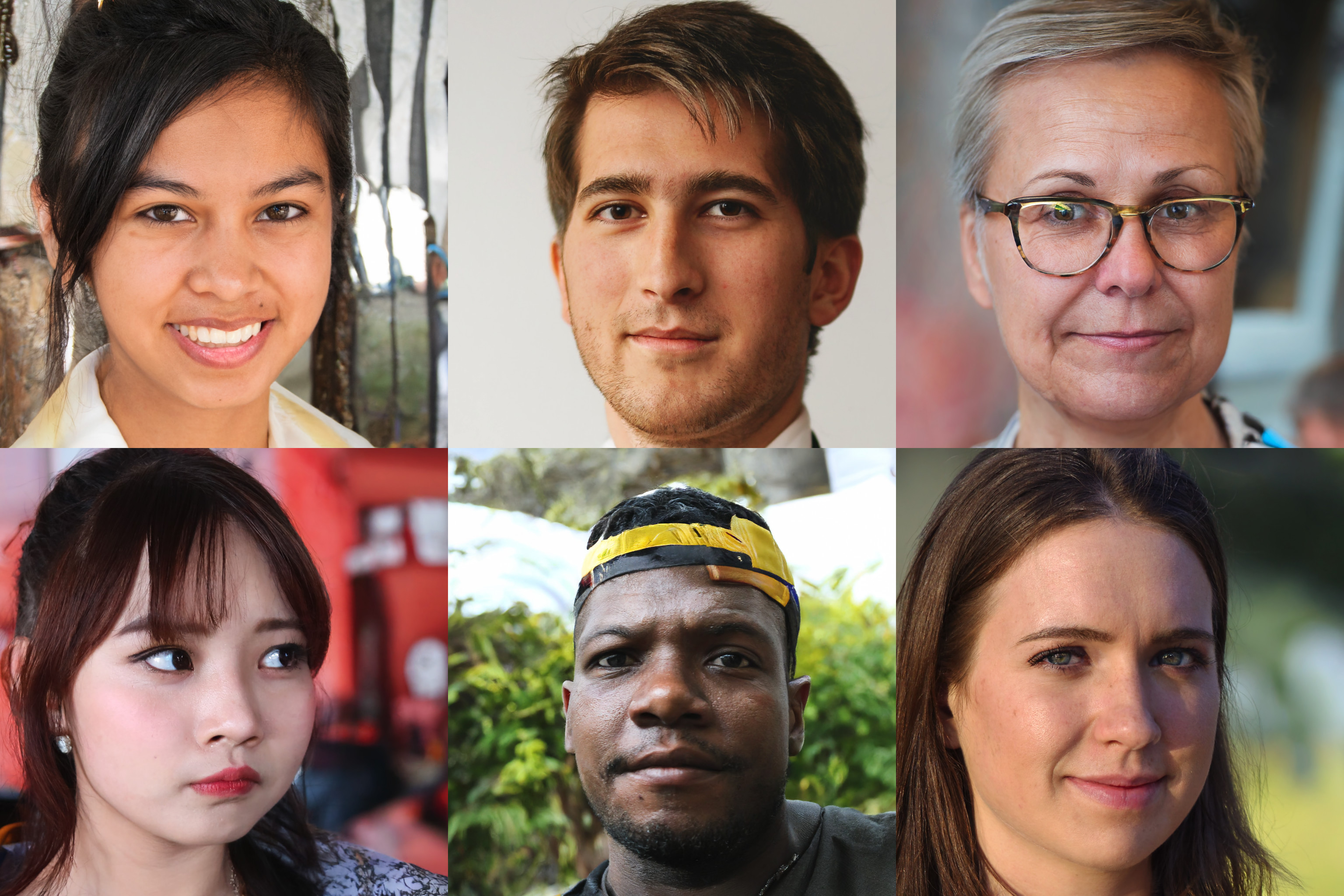}
    \caption{Samples from our 85M-parameter FFHQ-$1024^2$ model. Best viewed zoomed in.}
    \vspace{-7mm}
    \label{fig:ffhq}
\end{figure}

Previous works use self-conditioning \cite{jabri2023rin}, multi-scale architectures \cite{gu2023matryoshka}, or multi-scale losses \cite{hoogeboom2023simple} to enable synthesis at high resolutions. Our model does not require such tricks (though we expect them to further increase quality), and we train without them, with the exception of adapting the SNR at each step according to the increase in the images' redundancy \cite{hoogeboom2023simple}.
Our model generates high-quality, globally coherent samples (see \cref{fig:ffhq}) that utilize the high resolution to produce sharp pictures with fine details.

We benchmark our models against state-of-the-at counterparts in \cref{tab:ffhq} for a quantitative comparison.
We find that our model substantially outperforms this baseline both quantitatively and qualitatively (see \cref{fig:ffhq_uncurated} and \cref{fig:ffhq_curated} for samples from our method and competing methods). Notably, our model excels in generating faces with symmetric features, while the only other diffusion model, NCSN++, exhibits noticeable asymmetry. Moreover, \modelname{} effectively leverages the available resolution, producing sharp and finely detailed images, a notable improvement over the NCSN++ model, which often yields blurry samples, and also other competing methods.
We find that our model is competitive regarding FID with high-resolution transformer GANs such as HiT \cite{zhao2021improved} or StyleSwin \cite{zhang2022styleswin}, but does not reach the same FID as state-of-the-art GANs such as StyleGAN-XL \cite{sauer2022styleganxl}.
We evaluate with DINOv2-based metrics also, as FID is known to be flawed for evaluating FFHQ generation \cite{kynkaanniemi2023fidimagenetclasses} and to advantage GAN-generated samples \cite{stein2023exposing}. Our model sets a new state-of-the-art for DINOv2-based Fr\'echet and Kernel distances, metrics which correlate better with human preference than their Inception counterparts \cite{stein2023exposing}.

\begin{table}[tb]
    \centering
    \caption{Comparison of our results on FFHQ 1024 $\times$ 1024 to other models in the literature. 50k samples are used for FID computation unless specified otherwise.\textsuperscript{\ref{footnote:ncsn}} FD\textsubscript{D2} and KD\textsubscript{D2} denote Fr\'echet and Kernel DINOv2 distances respectively.}
    \begin{adjustbox}{max width=\columnwidth}
    \newcommand{\ffhqmethodtableentry}[9]{#1 & #9 & #2 & #3 & #4}
    \begin{tabular}{lcccc}
        \toprule
        \textbf{Method} & \textbf{Params} & \textbf{FID}\lowerisbetter & \textbf{FD\textsubscript{D2}}\lowerisbetter & \textbf{KD\textsubscript{D2}}\lowerisbetter \\
        \midrule
        \multicolumn{2}{l}{\textit{Diffusion Models (5k samples)}} \\
        \ffhqmethodtableentry{NCSN++ \cite{song2021scorebased}}{53.52}{608}{1.879}{0.839}{0.272}{1.104}{0.574}{106M} \\
        \ffhqmethodtableentry{\modelname{}-85M (\textbf{Ours})}{{\color{white}0}\textbf{8.48}}{\textbf{177}}{\textbf{0.348}}{0.772}{0.704}{1.028}{0.940}{{\color{white}0}85M} \\
        \midrule
        \multicolumn{2}{l}{\textit{Diffusion Models}} \\
        \ffhqmethodtableentry{\modelname{}-85M (\textbf{Ours})}{{\color{white}0}5.23}{\textbf{149}}{\textbf{0.354}}{0.769}{0.691}{1.020}{0.928}{{\color{white}0}85M} \\
        \arrayrulecolor{black!50} \midrule \arrayrulecolor{black}
        \multicolumn{2}{l}{{\color{black}\textit{Generative Adversarial Networks}}} \\
        \ffhqmethodtableentry{{\color{black}HiT-B \cite{zhao2021improved}}}{{\color{black}{\color{white}0}6.37}}{\color{black}-}{\color{black}-}{}{}{}{}{117M} \\
        \ffhqmethodtableentry{{\color{black}StyleSwin \cite{zhang2022styleswin}}}{{\color{black}{\color{white}0}5.07}}{\color{black}360}{\color{black}0.946}{0.802}{0.614}{1.236}{0.929}{{\color{white}0}41M} \\
        \ffhqmethodtableentry{{\color{black}StyleGAN2 \cite{karras2019stylegan2}}}{{\color{black}{\color{white}0}\underline{2.70}}}{\color{black}253}{\color{black}0.578}{0.796}{0.683}{1.115}{0.944}{{\color{white}0}30M} \\
        \ffhqmethodtableentry{{\color{black}StyleGAN3-T \cite{karras2021stylegan3}}}{{\color{black}{\color{white}0}2.79}}{\color{black}\underline{249}}{\color{black}\underline{0.575}}{0.762}{0.690}{0.996}{0.942}{{\color{white}0}22M} \\
        \ffhqmethodtableentry{{\color{black}StyleGAN3-R \cite{karras2021stylegan3}}}{{\color{black}{\color{white}0}3.07}}{\color{black}273}{\color{black}0.651}{0.752}{0.696}{0.979}{0.941}{{\color{white}0}16M} \\
        \ffhqmethodtableentry{{\color{black}StyleGAN-XL \cite{sauer2022styleganxl}}}{{\color{black}{\color{white}0}\textbf{2.02}}}{\color{black}270}{\color{black}0.644}{0.787}{0.680}{1.047}{0.946}{{\color{white}0}71M} \\
        \bottomrule
    \end{tabular}
    \end{adjustbox}
    \label{tab:ffhq}
\end{table}
\addtocounter{footnote}{1}\interfootnotelinepenalty=10000\footnotetext{\label{footnote:ncsn}We compare to NCSN++ on FID@5k due to its sampling cost, which for FID@50k would be similar to training our model.}

\subsection{Large-Scale ImageNet Image Synthesis}\label{sec:large_scale_imagenet}
Earlier experiments (see \cref{sec:high_res_image_synthesis}) show \modelname{}'s sample fidelity at high resolutions. To evaluate capabilities at scale, we train a class-conditional pixel-space ImageNet-$256^2$ model.
This 557M parameter model is smaller than many state-of-the-art models, and has not been hyperparameter-tuned.
As in our high-resolution experiments, we refrain from applying non-standard training tricks or diffusion modifications, and, consistent with \cite{hoogeboom2023simple}, we compare results without the application of classifier-free guidance, emphasizing an out-of-the-box comparison.

\begin{figure}[htb]
    \centering
    \includegraphics[width=\columnwidth]{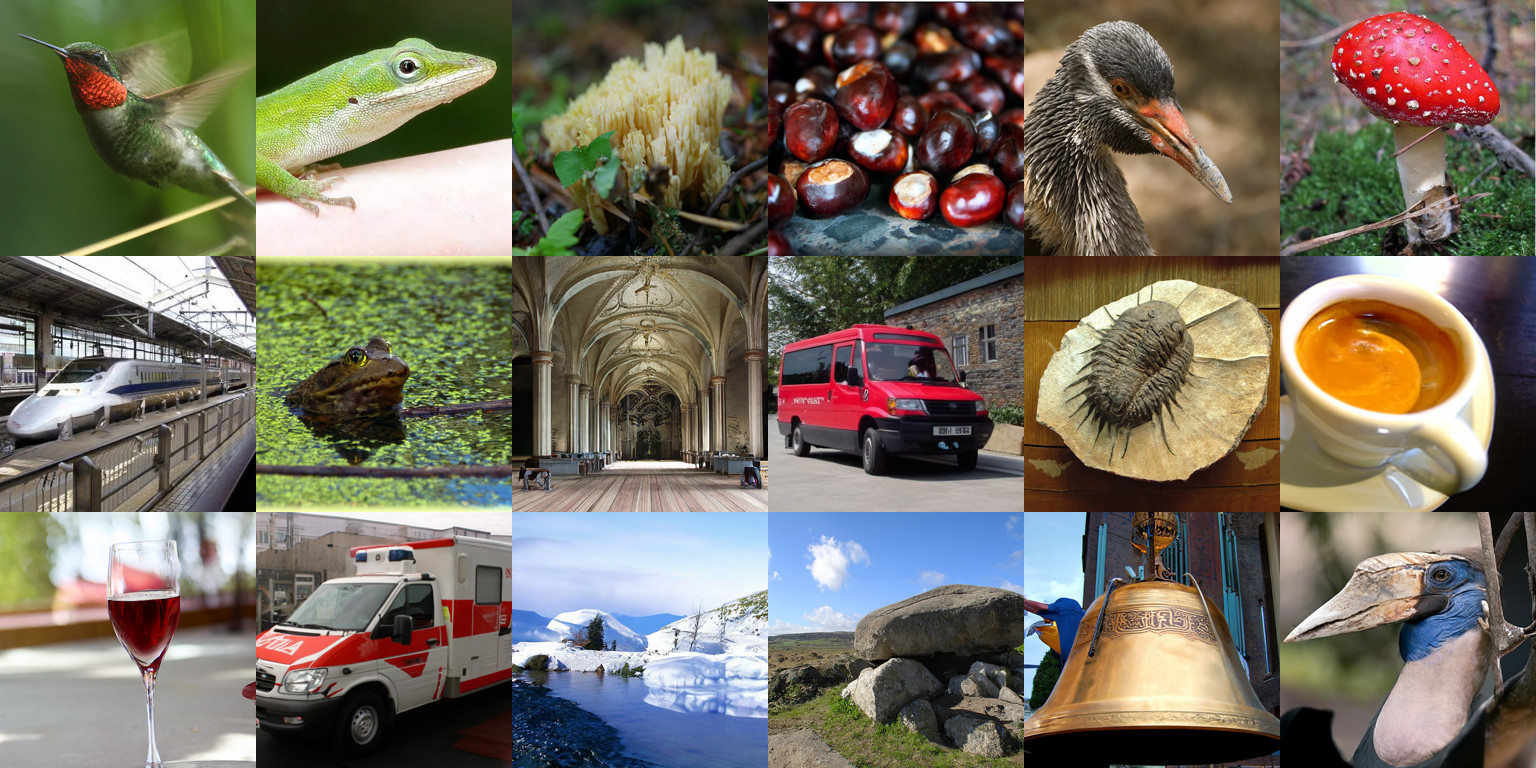}
    \caption{Samples from our class-conditional 557M-parameter ImageNet-$256^2$ model without CFG.}
    \vspace{-3mm}
    \label{fig:imagenet}
\end{figure}

We show samples in \cref{fig:imagenet} and compare quantitatively with state-of-the-art diffusion models in \cref{tab:imagenet}. We find that qualitatively our model can generate high-fidelity samples on this task. Compared to the baseline model DiT, our model achieves a substantially lower FID and higher IS despite operating on pixel-space instead of lower-resolution latents. Compared to other single-stage pixel-space diffusion models, our model outperforms simple U-Net-based models such as ADM but is outperformed by models that use self-conditioning during sampling (RIN) or are substantially larger (simple diffusion, VDM++).

\begin{table}[htb]
    \centering
    \caption{Comparison of our results on ImageNet-$256^2$ to other models in the literature. Following \cite{hoogeboom2023simple}, we report results without classifier-free guidance. Besides FID@50k and IS@50k, we also report trainable parameter count, samples seen (training iterations times batch size), and sampling steps.}
    \begin{adjustbox}{max width=\columnwidth}
    \begin{tabular}{l@{}c@{\hskip \tabcolsep}c@{\hskip \tabcolsep}c@{\hskip \tabcolsep}c@{\hskip \tabcolsep}c}
        \toprule
        \textbf{Method} & \textbf{Params} & \textbf{It.$\times$BS} & \textbf{Steps} & \textbf{FID}\lowerisbetter & \textbf{IS}\higherisbetter \\
        \midrule
        \multicolumn{5}{l}{\textit{Latent Diffusion Models}} \\
        LDM-4 \cite{rombach2022ldm} & 400M+VAE & 214M & 250 & 10.56 & 209.5 \\
        DiT-XL/2 \cite{peebles2023dit} & 675M+VAE & 1.8B & 250 & {\color{white}0}9.62 & 121.5 \\
        U-ViT-H/2 \cite{bao2022uvit} & 501M+VAE & 512M & 50$\cdot$2 & {\color{white}0}6.58 & - \\
        MDT-XL/2 \cite{gao2023mdt} & 676M+VAE & 1.7B & 250 & {\color{white}0}6.23 & 143.0 \\
        MaskDiT/2 \cite{zheng2023maskdit} & 736M+VAE & 2B & 40$\cdot$2 & {\color{white}0}5.69 & 178.0 \\
        \arrayrulecolor{black!50} \midrule \arrayrulecolor{black}
        \multicolumn{5}{l}{\textit{Single-Stage Pixel-Space Diffusion Models}} \\
        iDDPM \cite{nichol2021iddpm} & - & - & 250 & 32.50 & - \\
        ADM \cite{dhariwal2021diffusion} & 554M & 507M & 1000 & 10.94 & 101.0\\
        RIN \cite{jabri2023rin} & 410M & 614M & 1000 & {\color{white}0}4.51 & 161.0 \\
        simple diffusion \cite{hoogeboom2023simple} & 2B & 1B & 512 & {\color{white}0}2.77 & 211.8 \\
        VDM++ \cite{kingma2023vdmpp} & 2B & - & 256$\cdot$2 & {\color{white}0}2.40 & 225.3 \\
        \modelname{} (\textbf{Ours}) & 557M & 742M & 50$\cdot$2 & {\color{white}0}6.92 & 135.2 \\
        \bottomrule
    \end{tabular}
    \end{adjustbox}
    \vspace{-5mm}
    \label{tab:imagenet}
\end{table}

\section{Conclusion}\label{sec:conclusion}
This work presents \modelname{}, a hierarchical pure transformer backbone for diffusion image synthesis which scales to high resolutions more efficiently than previous transformer-based backbones. It adapts to the target resolution, processing local phenomena at high resolutions and global phenomena at low resolutions. Its computational complexity at higher resolutions scales with $\mathcal{O}(n)$ instead of $\mathcal{O}(n^2)$, bridging the gap between the scalability of transformer models and the efficiency of U-Nets. We demonstrate megapixel-scale pixel-space synthesis without tricks such as self-conditioning or multiresolution architectures, whilst staying competitive with other transformer diffusion backbones even at small resolutions, both in fairly matched pixel-space settings, and when compared to transformers in latent-space.

\section{Future Work}\label{sec:future_work}
\modelname{} provides a basis for further research into efficient high-resolution image synthesis. While we only focus on unconditional and class-conditional image synthesis, \modelname{} is likely well-suited to enhancing efficiency and performance in other generative tasks like super-resolution, text-to-image generation and other modalities such as audio and video, especially with architecture scaling.
This work studied \modelname{} in the context of pixel-space diffusion models but future works could investigate applying \modelname{} in a latent diffusion setup to increase efficiency further and achieve multi-megapixel image resolutions, or apply orthogonal tricks such as self-conditioning \cite{jabri2023rin} or progressive training \cite{sauer2022styleganxl} to improve the quality of generated samples further.

Our large-scale ImageNet experiment (see \cref{sec:large_scale_imagenet}) shows promise, competing with many state-of-the-art architectures. Future work could realize the potential of \modelname{} with hyperparameter tuning, architecture scaling, and recent practices \cite{karras2023edm2}.

Our architecture with local attention blocks could enable efficient diffusion superresolution and diffusion VAE feature decoding models: if all levels are set to perform local attention only (global attention blocks should not be necessary as the global structure is already present in the samples for these applications), one can train efficient transformer-based models that can scale to arbitrary resolutions.

\section*{Impact Statement}
This work aims to improve the capabilities of diffusion models by enabling the training of high-resolution pixel-space transformer-based diffusion models. While many other high-resolution diffusion models exist already, the majority do not operate in pixel-space. Operating in pixel-space potentially enables substantially higher-quality image editing and controllable generation capabilities as downstream tasks.
Especially in the context of image editing, capable image synthesis models such as Stable Diffusion \cite{rombach2022ldm} have been found to carry risks of generating harmful or deceptive content. In general, progress in high-resolution image synthesis contributes to the production of believable disinformation and could worsen society's ability to trust the authenticity of content. Whilst our method improves the efficiency of transformer-based diffusion models, it remains the case that training and inferencing of diffusion models is energy-intensive, potentially contributing to wider issues such as climate change.

\section*{Acknowledgements}
We thank Jack Gallagher for extensive help with early conceptual iterations and baseline implementation; Tao Hu for his extensive input and guidance, and \textit{uptightmoose} for their input during the paper writing process. We also thank the reviewers for their useful suggestions.
We gratefully acknowledge Jenia Jitsev for his advice on scaling law experiments and support for the final set of experiments in the revision, and LAION for access to compute budgets granted by Gauss Centre for Supercomputing e.V. and by the John von Neumann Institute for Computing (NIC) on the supercomputers JUWELS Booster and JURECA at Jülich Supercomputing Centre (JSC).
ES gratefully acknowledges Stability AI for resources to conduct experiments.

\bibliography{main}

@String(CVPR= {IEEE Conf. Comput. Vis. Pattern Recog.})

@String(ICCV= {Int. Conf. Comput. Vis.})

@String(ICLR = {Int. Conf. Learn. Represent.})

@String(CVPR  = {CVPR})

@String(ICCV  = {ICCV})

@String(ICLR  = {ICLR})

@inproceedings{nawrot2022hourglass,
      title = {{H}ierarchical {T}ransformers {A}re {M}ore {E}fficient {L}anguage {M}odels},
      author = "Nawrot, Piotr  and
      Tworkowski, Szymon  and
      Tyrolski, Micha{\l}  and
      Kaiser, Lukasz  and
      Wu, Yuhuai  and
      Szegedy, Christian  and
      Michalewski, Henryk",
      booktitle = "Findings of the Association for Computational Linguistics: NAACL 2022",
      month = JUL,
      year = "2022",
}

@inproceedings{peebles2023dit,
      author    = {Peebles, William and Xie, Saining},
      title     = {{S}calable {D}iffusion {M}odels with {T}ransformers},
      booktitle = {IEEE/CVF International Conference on Computer Vision (ICCV)},
      month     = {October},
      year      = {2023},
}

@inproceedings{hassani2023neighborhood,
      author    = {Hassani, Ali and Walton, Steven and Li, Jiachen and Li, Shen and Shi, Humphrey},
      title     = {{N}eighborhood {A}ttention {T}ransformer},
      booktitle = {IEEE/CVF Conference on Computer Vision and Pattern Recognition (CVPR)},
      month     = {June},
      year      = {2023},
}

@inproceedings{liu2021swin,
      author={Liu, Ze and Lin, Yutong and Cao, Yue and Hu, Han and Wei, Yixuan and Zhang, Zheng and Lin, Stephen and Guo, Baining},
      booktitle={IEEE/CVF International Conference on Computer Vision (ICCV)}, 
      title={{S}win {T}ransformer: {H}ierarchical {V}ision {T}ransformer using {S}hifted {W}indows}, 
      year={2021},
}

@inproceedings{liu2022swin,
      author={Liu, Ze and Hu, Han and Lin, Yutong and Yao, Zhuliang and Xie, Zhenda and Wei, Yixuan and Ning, Jia and Cao, Yue and Zhang, Zheng and Dong, Li and Wei, Furu and Guo, Baining},
      booktitle={IEEE/CVF Conference on Computer Vision and Pattern Recognition (CVPR)}, 
      title={{S}win {T}ransformer {V}2: {S}caling {U}p {C}apacity and {R}esolution}, 
      year={2022},
}

@inproceedings{wang2021uformer,
      author={Wang, Zhendong and Cun, Xiaodong and Bao, Jianmin and Zhou, Wengang and Liu, Jianzhuang and Li, Houqiang},
      booktitle={IEEE/CVF Conference on Computer Vision and Pattern Recognition (CVPR)}, 
      title={{U}former: {A} {G}eneral {U}-{S}haped {T}ransformer for {I}mage {R}estoration}, 
      year={2022},
}

@inproceedings{zamir2022restormer,
      title={{R}estormer: {E}fficient {T}ransformer for {H}igh-{R}esolution {I}mage {R}estoration}, 
    author={Syed Waqas Zamir and Aditya Arora and Salman Khan and Munawar Hayat 
            and Fahad Shahbaz Khan and Ming-Hsuan Yang},
      booktitle={2022 IEEE/CVF Conference on Computer Vision and Pattern Recognition (CVPR)}, 
      year={2022}
}

@misc{su2022roformer,
      title={{R}o{F}ormer: {E}nhanced {T}ransformer with {R}otary {P}osition {E}mbedding}, 
      author={Jianlin Su and Yu Lu and Shengfeng Pan and Ahmed Murtadha and Bo Wen and Yunfeng Liu},
      year={2022},
      eprint={2104.09864},
      archivePrefix={arXiv},
      primaryClass={cs.CL}
}

@inproceedings{zhang2019root,
      address = "Vancouver, Canada",
      author = "Zhang, Biao and Sennrich, Rico",
      booktitle = "Advances in Neural Information Processing Systems 32",
      title = "{Root Mean Square Layer Normalization}",
      year = "2019"
}

@inproceedings{karras2022elucidating,
      author    = {Tero Karras and Miika Aittala and Timo Aila and Samuli Laine},
      title     = {{E}lucidating the {D}esign {S}pace of {D}iffusion-{B}ased {G}enerative {M}odels},
      booktitle = {Conference on Neural Information Processing Systems (NeurIPS)},
      year      = {2022}
}

@misc{lu2023dpmsolver,
      title={{D}{P}{M}-{S}olver++: {F}ast {S}olver for {G}uided {S}ampling of {D}iffusion {P}robabilistic {M}odels}, 
      author={Cheng Lu and Yuhao Zhou and Fan Bao and Jianfei Chen and Chongxuan Li and Jun Zhu},
      year={2023},
      eprint={2211.01095},
      archivePrefix={arXiv},
      primaryClass={cs.LG}
}

@inproceedings{hang2023minsnr,
      author    = {Hang, Tiankai and Gu, Shuyang and Li, Chen and Bao, Jianmin and Chen, Dong and Hu, Han and Geng, Xin and Guo, Baining},
      title     = {{E}fficient {D}iffusion {T}raining via {M}in-{S}{N}{R} {W}eighting {S}trategy},
      booktitle = {IEEE/CVF International Conference on Computer Vision (ICCV)},
      month     = {October},
      year      = {2023},
}

@inproceedings{hoogeboom2023simple,
      author = {Hoogeboom, Emiel and Heek, Jonathan and Salimans, Tim},
      title = {{S}imple {D}iffusion: {E}nd-to-{E}nd {D}iffusion for {H}igh {R}esolution {I}mages},
      year = {2023},
      publisher = {JMLR.org},
      booktitle = {International Conference on Machine Learning (ICML)},
      articleno = {537},
      numpages = {20},
      location = {Honolulu, Hawaii, USA},
}

@inproceedings{rombach2022ldm,
      author    = {Rombach, Robin and Blattmann, Andreas and Lorenz, Dominik and Esser, Patrick and Ommer, Bj\"orn},
      title     = {{H}igh-{R}esolution {I}mage {S}ynthesis {W}ith {L}atent {D}iffusion {M}odels},
      booktitle = {IEEE/CVF Conference on Computer Vision and Pattern Recognition (CVPR)},
      month     = {June},
      year      = {2022},
}

@misc{ramesh2022dalle2,
      title={{H}ierarchical {T}ext-{C}onditional {I}mage {G}eneration with {C}{L}{I}{P} {L}atents}, 
      author={Aditya Ramesh and Prafulla Dhariwal and Alex Nichol and Casey Chu and Mark Chen},
      year={2022},
      eprint={2204.06125},
      archivePrefix={arXiv},
      primaryClass={cs.CV}
}

@inproceedings{dhariwal2021diffusion,
      title={{D}iffusion {M}odels {B}eat {{G}{A}{N}}s on {I}mage {S}ynthesis},
      author={Prafulla Dhariwal and Alexander Quinn Nichol},
      booktitle={Conference on Neural Information Processing Systems (NeurIPS)},
      year={2021},
}

@misc{zheng2023maskdit,
      title={{F}ast {T}raining of {D}iffusion {M}odels with {M}asked {T}ransformers}, 
      author={Hongkai Zheng and Weili Nie and Arash Vahdat and Anima Anandkumar},
      year={2023},
      eprint={2306.09305},
      archivePrefix={arXiv},
      primaryClass={cs.CV}
}

@inproceedings{gao2023mdt,
      title={{M}asked {D}iffusion {T}ransformer is a {S}trong {I}mage {S}ynthesizer}, 
      author={Shanghua Gao and Pan Zhou and Ming-Ming Cheng and Shuicheng Yan},
      year={2023},
      booktitle={IEEE/CVF International Conference on Computer Vision (ICCV)},
      month={October}
}

@inproceedings{bao2022uvit,
  title={{A}ll are {{W}}orth {{W}}ords: {{A}} {{V}}i{{T}} {{B}}ackbone for {{D}}iffusion {{M}}odels},
  author={Bao, Fan and Nie, Shen and Xue, Kaiwen and Cao, Yue and Li, Chongxuan and Su, Hang and Zhu, Jun},
  booktitle = {IEEE/CVF Conference on Computer Vision and Pattern Recognition (CVPR)},
  year={2023}
}

@inproceedings{vaswani2017transformer,
 author = {Vaswani, Ashish and Shazeer, Noam and Parmar, Niki and Uszkoreit, Jakob and Jones, Llion and Gomez, Aidan N and Kaiser, \L ukasz and Polosukhin, Illia},
 booktitle = {Conference on Neural Information Processing Systems (NeurIPS)},
 pages = {},
 title = {{A}ttention is {A}ll you {N}eed},
 year = {2017}
}

@misc{jing2023udittts,
      title={{U}-{D}i{T} {T}{T}{S}: {U}-{D}iffusion {V}ision {T}ransformer for {T}ext-to-{S}peech}, 
      author={Xin Jing and Yi Chang and Zijiang Yang and Jiangjian Xie and Andreas Triantafyllopoulos and Bjoern W. Schuller},
      year={2023},
      eprint={2305.13195},
      archivePrefix={arXiv},
      primaryClass={cs.SD}
}

@inproceedings{bao2023unidiffusion,
    author = {Bao, Fan and Nie, Shen and Xue, Kaiwen and Li, Chongxuan and Pu, Shi and Wang, Yaole and Yue, Gang and Cao, Yue and Su, Hang and Zhu, Jun},
    title = {{O}ne {T}ransformer {F}its {A}ll {D}istributions in {M}ulti-{M}odal {D}iffusion at {S}cale},
    year = {2023},
    publisher = {JMLR.org},
    booktitle = {International Conference on Machine Learning (ICML)},
    articleno = {72},
    numpages = {26},
    location = {Honolulu, Hawaii, USA},
}

@inproceedings{dehghani2023scalingvit,
    author = {Dehghani, Mostafa and Djolonga, Josip and Mustafa, Basil and Padlewski, Piotr and Heek, Jonathan and Gilmer, Justin and Steiner, Andreas and Caron, Mathilde and Geirhos, Robert and Alabdulmohsin, Ibrahim and Jenatton, Rodolphe and Beyer, Lucas and Tschannen, Michael and Arnab, Anurag and Wang, Xiao and Riquelme, Carlos and Minderer, Matthias and Puigcerver, Joan and Evci, Utku and Kumar, Manoj and Van Steenkiste, Sjoerd and Elsayed, Gamaleldin F. and Mahendran, Aravindh and Yu, Fisher and Oliver, Avital and Huot, Fantine and Bastings, Jasmijn and Collier, Mark Patrick and Gritsenko, Alexey A. and Birodkar, Vighnesh and Vasconcelos, Cristina and Tay, Yi and Mensink, Thomas and Kolesnikov, Alexander and Paveti\'{c}, Filip and Tran, Dustin and Kipf, Thomas and Lu\v{c}i\'{c}, Mario and Zhai, Xiaohua and Keysers, Daniel and Harmsen, Jeremiah and Houlsby, Neil},
    title = {{S}caling {V}ision {T}ransformers to 22 {B}illion {P}arameters},
    year = {2023},
    booktitle = {International Conference on Machine Learning (ICML)},
    publisher = {JMLR.org},
    articleno = {296},
    numpages = {33},
    location = {Honolulu, Hawaii, USA},
}

@inproceedings{ronneberger2015unet,
    author = {Ronneberger, Olaf and Fischer, Philipp and Brox, Thomas},
    booktitle = {Medical Image Computing and Computer-Assisted Intervention (MICCAI)},
    title = {{U}-{N}et: {C}onvolutional {N}etworks for {B}iomedical {I}mage {S}egmentation},
    year = {2015}
}

@inproceedings{ho2020ddpm,
    author = {Ho, Jonathan and Jain, Ajay and Abbeel, Pieter},
    title = {{D}enoising {D}iffusion {P}robabilistic {M}odels},
    year = {2020},
    booktitle = {Conference on Neural Information Processing Systems (NeurIPS)},
}

@misc{stein2023exposing,
    title={{E}xposing flaws of generative model evaluation metrics and their unfair treatment of diffusion models}, 
    author={George Stein and Jesse C. Cresswell and Rasa Hosseinzadeh and Yi Sui and Brendan Leigh Ross and Valentin Villecroze and Zhaoyan Liu and Anthony L. Caterini and J. Eric T. Taylor and Gabriel Loaiza-Ganem},
    year={2023},
}

@misc{cao2022exploring,
    title={{E}xploring {V}ision {T}ransformers as {D}iffusion {L}earners}, 
    author={He Cao and Jianan Wang and Tianhe Ren and Xianbiao Qi and Yihao Chen and Yuan Yao and Lei Zhang},
    year={2022},
    eprint={2212.13771},
    archivePrefix={arXiv},
    primaryClass={cs.CV}
}

@INPROCEEDINGS{jia2009imagenet,
    author={Deng, Jia and Dong, Wei and Socher, Richard and Li, Li-Jia and Kai Li and Li Fei-Fei},
    booktitle={IEEE Conference on Computer Vision and Pattern Recognition (CVPR)}, 
    title={{I}mage{N}et: {A} large-scale hierarchical image database}, 
    year={2009},
}

@inproceedings{loshchilov2018adamw,
    title={{D}ecoupled {W}eight {D}ecay {R}egularization},
    author={Ilya Loshchilov and Frank Hutter},
    booktitle={International Conference on Learning Representations (ICLR)},
    year={2019},
}

@inproceedings{ho2021classifierfree,
    title={{C}lassifier-{F}ree {D}iffusion {G}uidance},
    author={Jonathan Ho and Tim Salimans},
    booktitle={NeurIPS 2021 Workshop on Deep Generative Models and Downstream Applications},
    year={2021},
}

@inproceedings{heusel2017fid,
    author = {Heusel, Martin and Ramsauer, Hubert and Unterthiner, Thomas and Nessler, Bernhard and Hochreiter, Sepp},
    booktitle = {Conference on Neural Information Processing Systems (NeurIPS)},
    editor = {I. Guyon and U. Von Luxburg and S. Bengio and H. Wallach and R. Fergus and S. Vishwanathan and R. Garnett},
    title = {{G}{A}{N}s {T}rained by a {T}wo {T}ime-{S}cale {U}pdate {R}ule {C}onverge to a {L}ocal {N}ash {E}quilibrium},
    year = {2017}
}

@misc{flash-cosine-sim-attention,
    Author = {Phil Wang},
    Year = {2022},
    url = {https://github.com/lucidrains/flash-cosine-sim-attention/tree/6f17f29a979a8bcab2479c65b7740523},
    Title = {Flash {C}osine {S}imilarity {A}ttention}
}

@InProceedings{zhang2022styleswin,
    author    = {Zhang, Bowen and Gu, Shuyang and Zhang, Bo and Bao, Jianmin and Chen, Dong and Wen, Fang and Wang, Yong and Guo, Baining},
    title     = {{S}tyle{S}win: {T}ransformer-{B}ased {G}{A}{N} for {H}igh-{R}esolution {I}mage {G}eneration},
    booktitle = {Proceedings of the IEEE/CVF Conference on Computer Vision and Pattern Recognition (CVPR)},
    month     = {June},
    year      = {2022},
    pages     = {11304-11314}
}

@misc{shazeer2020glu,
      title={{G}{L}{U} {V}ariants {I}mprove {T}ransformer}, 
      author={Noam Shazeer},
      year={2020},
      eprint={2002.05202},
      archivePrefix={arXiv},
      primaryClass={cs.LG}
}

@inproceedings{henry2020qknorm,
    title = "Query-Key Normalization for Transformers",
    author = "Henry, Alex  and
      Dachapally, Prudhvi Raj  and
      Pawar, Shubham Shantaram  and
      Chen, Yuxuan",
    editor = "Cohn, Trevor  and
      He, Yulan  and
      Liu, Yang",
    booktitle = "Findings of the Association for Computational Linguistics: EMNLP 2020",
    month = nov,
    year = "2020",
}

@misc{swin-v2-attention-implementation,
    Author = {Ze Liu and Han Hu and Yutong Lin and Zhuliang Yao and Zhenda Xie and Yixuan Wei and Jia Ning and Yue Cao and Zheng Zhang and Li Dong and Furu Wei and Baining Guo},
    Year = {2022},
    url = {https://github.com/microsoft/Swin-Transformer/blob/2cb103f2de145ff43bb9f6fc2ae8800c24/models/swin\_transformer\_v2.py\#L156},
    Title = {{S}{W}in {T}ransformer v2}
}

@INPROCEEDINGS {shi2016pixelshuffle,
author = {W. Shi and J. Caballero and F. Huszar and J. Totz and A. P. Aitken and R. Bishop and D. Rueckert and Z. Wang},
booktitle = {IEEE/CVF Conference on Computer Vision and Pattern Recognition (CVPR)},
title = {{R}eal-{T}ime {S}ingle {I}mage and {V}ideo {S}uper-{R}esolution {U}sing an {E}fficient {S}ub-{P}ixel {C}onvolutional {N}eural {N}etwork},
year = {2016},
month = {jun}
}

@techreport{touvron2023llama,
      title={{L}{L}a{M}{A}: {O}pen and {E}fficient {F}oundation {L}anguage {M}odels}, 
      author={Hugo Touvron and Thibaut Lavril and Gautier Izacard and Xavier Martinet and Marie-Anne Lachaux and Timothée Lacroix and Baptiste Rozière and Naman Goyal and Eric Hambro and Faisal Azhar and Aurelien Rodriguez and Armand Joulin and Edouard Grave and Guillaume Lample},
      year={2023},
      eprint={2302.13971},
      archivePrefix={arXiv},
      primaryClass={cs.CL}
}

@misc{ho2019axial,
    title = {{A}xial {A}ttention in {M}ultidimensional {T}ransformers},
    author = {Jonathan Ho and Nal Kalchbrenner and Dirk Weissenborn and Tim Salimans},
    year = {2019},
    archivePrefix = {arXiv}
}

@inproceedings{karras2020styleganada,
author = {Karras, Tero and Aittala, Miika and Hellsten, Janne and Laine, Samuli and Lehtinen, Jaakko and Aila, Timo},
title = {{T}raining {G}enerative {A}dversarial {N}etworks with {L}imited {D}ata},
year = {2020},
booktitle = {Conference on Neural Information Processing Systems (NeurIPS)},
articleno = {1015},
numpages = {11},
location = {Vancouver, BC, Canada},
}

@ARTICLE{karras2021stylegan,
author = {T. Karras and S. Laine and T. Aila},
journal = {IEEE Transactions on Pattern Analysis and Machine Intelligence (TPAMI)},
title = {{A} {S}tyle-{B}ased {G}enerator {A}rchitecture for {G}enerative {A}dversarial {N}etworks},
year = {2021},
publisher = {IEEE Computer Society},
address = {Los Alamitos, CA, USA},
}

@inproceedings{karras2021stylegan3,
  author = {Tero Karras and Miika Aittala and Samuli Laine and Erik H\"ark\"onen and Janne Hellsten and Jaakko Lehtinen and Timo Aila},
  title = {{A}lias-{F}ree {G}enerative {A}dversarial {N}etworks},
  booktitle = {Conference on Neural Information Processing Systems (NeurIPS)},
  year = {2021}
}

@inproceedings{sauer2022styleganxl,
author = {Sauer, Axel and Schwarz, Katja and Geiger, Andreas},
title = {{S}tyle{G}{A}{N}-{X}{L}: {S}caling {S}tyle{G}{A}{N} to {L}arge {D}iverse {D}atasets},
year = {2022},
publisher = {Association for Computing Machinery},
booktitle = {ACM SIGGRAPH 2022 Conference Proceedings},
location = {Vancouver, BC, Canada},
}

@misc{jabri2023rin,
      title={{S}calable {A}daptive {C}omputation for {I}terative {G}eneration}, 
      author={Allan Jabri and David Fleet and Ting Chen},
      year={2023},
      eprint={2212.11972},
      archivePrefix={arXiv},
      primaryClass={cs.LG}
}

@misc{gu2023matryoshka,
      title={{M}atryoshka {D}iffusion {M}odels}, 
      author={Jiatao Gu and Shuangfei Zhai and Yizhe Zhang and Josh Susskind and Navdeep Jaitly},
      year={2023},
      eprint={2310.15111},
      archivePrefix={arXiv},
      primaryClass={cs.CV}
}

@inproceedings{
  song2021scorebased,
  title={{S}core-{B}ased {G}enerative {M}odeling through {S}tochastic {D}ifferential {E}quations},
  author={Yang Song and Jascha Sohl-Dickstein and Diederik P Kingma and Abhishek Kumar and Stefano Ermon and Ben Poole},
  booktitle={International Conference on Learning Representations (ICLR)},
  year={2021},
}

@inproceedings{karras2019stylegan2,
  title     = {{A}nalyzing and {I}mproving the {I}mage {Q}uality of {{S}tyle{G}{A}{N}}},
  author    = {Tero Karras and Samuli Laine and Miika Aittala and Janne Hellsten and Jaakko Lehtinen and Timo Aila},
  booktitle = {IEEE/CVF Conference on Computer Vision and Pattern Recognition (CVPR)},
  year      = {2020}
}

@misc{kingma2023vdmpp,
      title={{U}nderstanding {D}iffusion {O}bjectives as the {E}{L}{B}{O} with {S}imple {D}ata {A}ugmentation}, 
      author={Diederik P. Kingma and Ruiqi Gao},
      year={2023},
      eprint={2303.00848},
      archivePrefix={arXiv},
      primaryClass={cs.LG}
}

@inproceedings{nichol2021iddpm,
  title={{I}mproved denoising diffusion probabilistic models},
  author={Nichol, Alexander Quinn and Dhariwal, Prafulla},
  booktitle={International Conference on Machine Learning (ICML)},
  year={2021},
  organization={PMLR}
}

@inproceedings{kynkaanniemi2023fidimagenetclasses,
title={{T}he {R}ole of {I}mage{N}et {C}lasses in {F}r\'echet {I}nception {D}istance},
author={Tuomas Kynk{\"a}{\"a}nniemi and Tero Karras and Miika Aittala and Timo Aila and Jaakko Lehtinen},
booktitle={International Conference on Learning Representations (ICLR)},
year={2023},
url={https://openreview.net/forum?id=4oXTQ6m_ws8}
}

@misc{li2022swinv2imagen,
      title={{S}winv2-{I}magen: {H}ierarchical {V}ision {T}ransformer {D}iffusion {M}odels for {T}ext-to-{I}mage {G}eneration}, 
      author={Ruijun Li and Weihua Li and Yi Yang and Hanyu Wei and Jianhua Jiang and Quan Bai},
      year={2022},
      eprint={2210.09549},
      archivePrefix={arXiv},
      primaryClass={cs.CV}
}

@misc{yang2022genvit,
      title={{Y}our {V}i{T} is {S}ecretly a {H}ybrid {D}iscriminative-{G}enerative {D}iffusion {M}odel}, 
      author={Xiulong Yang and Sheng-Min Shih and Yinlin Fu and Xiaoting Zhao and Shihao Ji},
      year={2022},
      eprint={2208.07791},
      archivePrefix={arXiv},
      primaryClass={cs.CV}
}

@misc{fischer2023boosting,
      title={{B}oosting {L}atent {D}iffusion with {F}low {M}atching}, 
      author={Johannes S. Fischer and Ming Gui and Pingchuan Ma and Nick Stracke and Stefan A. Baumann and Björn Ommer},
      year={2023},
      eprint={2312.07360},
      archivePrefix={arXiv},
      primaryClass={cs.CV}
}

@misc{ho2021cascaded,
  title={{C}ascaded {D}iffusion {M}odels for {H}igh {F}idelity {I}mage {G}eneration},
  author={Ho, Jonathan and Saharia, Chitwan and Chan, William and Fleet, David J and Norouzi, Mohammad and Salimans, Tim},
  journal={arXiv preprint arXiv:2106.15282},
  year={2021}
}

@techreport{betker2023dalle3,
  title = {{I}mproving {{{I}mage {G}eneration}} with {{{B}etter {C}aptions}}},
  author = {Betker, James and Goh, Gabriel and Jing, Li and Brooks, Tim and Wang, Jianfeng and Li, Linjie and Ouyang, Long and Zhuang, Juntang and Lee, Joyce and Guo, Yufei and Manassra, Wesam and Dhariwal, Prafulla and Chu, Casey and Jiao, Yunxin and Ramesh, Aditya},
  year={2023},
  langid = {english}
}

@misc{chen2023pixartalpha,
    title={{P}ix{A}rt-$\alpha$: {F}ast {T}raining of {D}iffusion {T}ransformer for {P}hotorealistic {T}ext-to-{I}mage {S}ynthesis}, 
    author={Junsong Chen and Jincheng Yu and Chongjian Ge and Lewei Yao and Enze Xie and Yue Wu and Zhongdao Wang and James Kwok and Ping Luo and Huchuan Lu and Zhenguo Li},
    year={2023},
    eprint={2310.00426},
    archivePrefix={arXiv},
    primaryClass={cs.CV}
}

@misc{chen2023gentron,
      title={{G}en{T}ron: {D}elving {D}eep into {D}iffusion {T}ransformers for {I}mage and {V}ideo {G}eneration}, 
      author={Shoufa Chen and Mengmeng Xu and Jiawei Ren and Yuren Cong and Sen He and Yanping Xie and Animesh Sinha and Ping Luo and Tao Xiang and Juan-Manuel Perez-Rua},
      year={2023},
      eprint={2312.04557},
      archivePrefix={arXiv},
      primaryClass={cs.CV}
}

@inproceedings{karras2023edm2,
      title={{A}nalyzing and {I}mproving the {T}raining {D}ynamics of {D}iffusion {M}odels}, 
      author={Tero Karras and Miika Aittala and Jaakko Lehtinen and Janne Hellsten and Timo Aila and Samuli Laine},
    booktitle={IEEE/CVF Conference on Computer Vision and Pattern Recognition (CVPR)},
    year={2023}
}

@inproceedings{
saharia2022photorealistic,
title={Photorealistic {T}ext-to-{I}mage {D}iffusion {M}odels with {D}eep {L}anguage {U}nderstanding},
author={Chitwan Saharia and William Chan and Saurabh Saxena and Lala Li and Jay Whang and Emily Denton and Seyed Kamyar Seyed Ghasemipour and Raphael Gontijo-Lopes and Burcu Karagol Ayan and Tim Salimans and Jonathan Ho and David J. Fleet and Mohammad Norouzi},
booktitle={Conference on Neural Information Processing Systems (NeurIPS)},
editor={Alice H. Oh and Alekh Agarwal and Danielle Belgrave and Kyunghyun Cho},
year={2022},
}

@misc{balaji2023ediffi,
      title={e{D}iff-{I}: {T}ext-to-{I}mage {D}iffusion {M}odels with an {E}nsemble of {E}xpert {D}enoisers}, 
      author={Yogesh Balaji and Seungjun Nah and Xun Huang and Arash Vahdat and Jiaming Song and Qinsheng Zhang and Karsten Kreis and Miika Aittala and Timo Aila and Samuli Laine and Bryan Catanzaro and Tero Karras and Ming-Yu Liu},
      year={2023},
      eprint={2211.01324},
      archivePrefix={arXiv},
      primaryClass={cs.CV}
}

@misc{dai2023emu,
      title={{E}mu: {E}nhancing {I}mage {G}eneration {M}odels {U}sing {P}hotogenic {N}eedles in a {H}aystack}, 
      author={Xiaoliang Dai and Ji Hou and Chih-Yao Ma and Sam Tsai and Jialiang Wang and Rui Wang and Peizhao Zhang and Simon Vandenhende and Xiaofang Wang and Abhimanyu Dubey and Matthew Yu and Abhishek Kadian and Filip Radenovic and Dhruv Mahajan and Kunpeng Li and Yue Zhao and Vladan Petrovic and Mitesh Kumar Singh and Simran Motwani and Yi Wen and Yiwen Song and Roshan Sumbaly and Vignesh Ramanathan and Zijian He and Peter Vajda and Devi Parikh},
      year={2023},
      eprint={2309.15807},
      archivePrefix={arXiv},
      primaryClass={cs.CV}
}

@inproceedings{kong2021diffwave,
title={{D}iff{W}ave: {A} {V}ersatile {D}iffusion {M}odel for {A}udio {S}ynthesis},
author={Zhifeng Kong and Wei Ping and Jiaji Huang and Kexin Zhao and Bryan Catanzaro},
booktitle={International Conference on Learning Representations (ICLR)},
year={2021},
}

@inproceedings{blattmann2023videoldm,
    title={Align your {L}atents: {H}igh-{R}esolution {V}ideo {S}ynthesis with {L}atent {D}iffusion {M}odels},
    author={Blattmann, Andreas and Rombach, Robin and Ling, Huan and Dockhorn, Tim and Kim, Seung Wook and Fidler, Sanja and Kreis, Karsten},
    booktitle={IEEE/CVF Conference on Computer Vision and Pattern Recognition (CVPR)},
    year={2023}
}

@inproceedings{zhao2021improved,
  title = {Improved {T}ransformer for {H}igh-{R}esolution {GANs}},
  author = {Long Zhao and Zizhao Zhang and Ting Chen and Dimitris Metaxas and Han Zhang},
  booktitle = {Conference on Neural Information Processing Systems (NeurIPS)},
  year = {2021}
}

@misc{yan2023diffusionwithoutattention,
      title={Diffusion {M}odels {W}ithout {A}ttention}, 
      author={Jing Nathan Yan and Jiatao Gu and Alexander M. Rush},
      year={2023},
}

@article{fedus2022switchtransformers,
  author  = {William Fedus and Barret Zoph and Noam Shazeer},
  title   = {Switch {T}ransformers: {S}caling to {T}rillion {P}arameter {M}odels with {S}imple and {E}fficient {S}parsity},
  journal = {Journal of Machine Learning Research (JMLR)},
  year    = {2022},
}

@article{chowdhery2023palm,
  author  = {Aakanksha Chowdhery and Sharan Narang and Jacob Devlin and Maarten Bosma and Gaurav Mishra and Adam Roberts and Paul Barham and Hyung Won Chung and Charles Sutton and Sebastian Gehrmann and Parker Schuh and Kensen Shi and Sasha Tsvyashchenko and Joshua Maynez and Abhishek Rao and Parker Barnes and Yi Tay and Noam Shazeer and Vinodkumar Prabhakaran and Emily Reif and Nan Du and Ben Hutchinson and Reiner Pope and James Bradbury and Jacob Austin and Michael Isard and Guy Gur-Ari and Pengcheng Yin and Toju Duke and Anselm Levskaya and Sanjay Ghemawat and Sunipa Dev and Henryk Michalewski and Xavier Garcia and Vedant Misra and Kevin Robinson and Liam Fedus and Denny Zhou and Daphne Ippolito and David Luan and Hyeontaek Lim and Barret Zoph and Alexander Spiridonov and Ryan Sepassi and David Dohan and Shivani Agrawal and Mark Omernick and Andrew M. Dai and Thanumalayan Sankaranarayana Pillai and Marie Pellat and Aitor Lewkowycz and Erica Moreira and Rewon Child and Oleksandr Polozov and Katherine Lee and Zongwei Zhou and Xuezhi Wang and Brennan Saeta and Mark Diaz and Orhan Firat and Michele Catasta and Jason Wei and Kathy Meier-Hellstern and Douglas Eck and Jeff Dean and Slav Petrov and Noah Fiedel},
  title   = {Pa{LM}: {S}caling {L}anguage {M}odeling with {P}athways},
  journal = {Journal of Machine Learning Research (JMLR)},
  year    = {2023},
}

@article{yu2022coca,
title={Co{C}a: {C}ontrastive {C}aptioners are {I}mage-{T}ext {F}oundation {M}odels},
author={Jiahui Yu and Zirui Wang and Vijay Vasudevan and Legg Yeung and Mojtaba Seyedhosseini and Yonghui Wu},
journal={Transactions on Machine Learning Research (TMLR)},
year={2022},
}

@inproceedings{zong2022codetr,
      title={{DETR}s with {C}ollaborative {H}ybrid {A}ssignments {T}raining},
      author={Zhuofan Zong and Guanglu Song and Yu Liu},
      year={2022},
      booktitle = {IEEE/CVF International Conference on Computer Vision (ICCV)},
}

@InProceedings{piergiovanni2023tubevit,
    author    = {Piergiovanni, AJ and Kuo, Weicheng and Angelova, Anelia},
    title     = {Rethinking {V}ideo {V}i{T}s: {S}parse {V}ideo {T}ubes for {J}oint {I}mage and {V}ideo {L}earning},
    booktitle = {IEEE/CVF Conference on Computer Vision and Pattern Recognition (CVPR)},
    year      = {2023},
}

@misc{zhang2022pushing,
      title={Pushing the {L}imits of {S}emi-{S}upervised {L}earning for {A}utomatic {S}peech {R}ecognition}, 
      author={Yu Zhang and James Qin and Daniel S. Park and Wei Han and Chung-Cheng Chiu and Ruoming Pang and Quoc V. Le and Yonghui Wu},
      year={2022},
}

@techreport{openai2023gpt4,
      title={{GPT}-4 {T}echnical {R}eport}, 
      author={OpenAI},
      year={2023},
}

@article{Saremi2013,
  title = {{H}ierarchical model of natural images and the origin of scale invariance},
  volume = {110},
  ISSN = {1091-6490},
  DOI = {10.1073/pnas.1222618110},
  number = {8},
  journal = {Proceedings of the National Academy of Sciences},
  publisher = {Proceedings of the National Academy of Sciences},
  author = {Saremi,  Saeed and Sejnowski,  Terrence J.},
  year = {2013},
  month = feb,
  pages = {3071–3076}
}

@InProceedings{huang2017adain,
    author = {Huang, Xun and Belongie, Serge},
    title = {Arbitrary {S}tyle {T}ransfer in {R}eal-{T}ime {W}ith {A}daptive {I}nstance {N}ormalization},
    booktitle = {IEEE/CVF International Conference on Computer Vision (ICCV)},
    month = {Oct},
    year = {2017}
}

@misc{yang2022feature,
      title={Feature Learning in Infinite-Width Neural Networks}, 
      author={Greg Yang and Edward J. Hu},
      year={2022},
      eprint={2011.14522},
      archivePrefix={arXiv},
      primaryClass={cs.LG}
}

@misc{yang2022tensor,
      title={Tensor Programs V: Tuning Large Neural Networks via Zero-Shot Hyperparameter Transfer}, 
      author={Greg Yang and Edward J. Hu and Igor Babuschkin and Szymon Sidor and Xiaodong Liu and David Farhi and Nick Ryder and Jakub Pachocki and Weizhu Chen and Jianfeng Gao},
      year={2022},
      eprint={2203.03466},
      archivePrefix={arXiv},
      primaryClass={cs.LG}
}

@misc{shonenkov2023deepfloydif,
    title={DeepFloyd IF},
    author={A Shonenkov and M Konstantinov and D Bakshandaeva and C Schuhmann and K Ivanova and N Klokova},
    year={2023},
}

@article{ba2016layer,
  title={Layer normalization},
  author={Ba, Jimmy Lei and Kiros, Jamie Ryan and Hinton, Geoffrey E},
  journal={arXiv preprint arXiv:1607.06450},
  year={2016}
}

@article{lu2022dpm,
  title={Dpm-solver: A fast ode solver for diffusion probabilistic model sampling in around 10 steps},
  author={Lu, Cheng and Zhou, Yuhao and Bao, Fan and Chen, Jianfei and Li, Chongxuan and Zhu, Jun},
  journal={Advances in Neural Information Processing Systems},
  volume={35},
  pages={5775--5787},
  year={2022}
}
\bibliographystyle{icml2024}

\newpage
\appendix


\clearpage

\newcommand{\ablationidminsnrfive}{\textbf{E2}}
\newcommand{\ablationidminsnrfour}{\textbf{E3}}
\newcommand{\ablationidskipadd}{\textbf{F1}}
\newcommand{\ablationidskipconcat}{\textbf{F2}}
\newcommand{\ablationidadaln}{\textbf{G}}
\newcommand{\ablationidadalngegluoutgate}{\textbf{H}}

\section{Computational Complexity of HDiT}\label{sec:computational_complexity}
    
\newcommand{\ops}[1]{\mathrm{ops}_\mathrm{#1}}
\newcommand{\mdim}[1]{d_\mathrm{#1}}
\newcommand{\resolution}[1]{\mathrm{res}_\mathrm{#1}}
\newcommand{\levels}[1]{\mathrm{levels}_\mathrm{#1}}
\newcommand{\ldepth}[1]{\mathrm{D}_\mathrm{#1}}

\newcommand{\globalattnops}{\ops{attn,global}}
\newcommand{\nattenops}{\ops{attn,neighborhood}}
\newcommand{\ffnops}{\ops{FFN}}
\newcommand{\ffncumops}{\ops{FFNs}}
\newcommand{\qdim}{\mdim{Q}}
\newcommand{\vdim}{\mdim{V}}
\newcommand{\modeldim}{\mdim{model}}
\newcommand{\hiddendim}{\mdim{hidden}}
\newcommand{\headdim}{\mdim{head}}
\newcommand{\minres}{\resolution{min}}
\newcommand{\maxgattnres}{\resolution{max,global}}
\newcommand{\totallevels}{\levels{total}}
\newcommand{\localattnlevels}{\levels{local}}
\newcommand{\globalattnlevels}{\levels{global}}
\newcommand{\outerdepth}{\ldepth{outer}}
\newcommand{\innerdepth}{\ldepth{inner}}

In a traditional vision transformer, including those for diffusion models \cite{peebles2023dit,bao2022uvit}, the asymptotic computational complexity with regard to image size is dominated by the self-attention mechanism, which scales as $\mathcal{O}(n^2d)$ with token/pixel count $n$ and embedding dimension $d$. The feedforward blocks and the attention projection heads, in turn, scale as $\mathcal{O}(nd^2)$.

For our Hourglass Diffusion Transformer architecture, we adjust the architecture for different target resolutions, similarly to previous approaches used with U-Nets \cite{ronneberger2015unet}. Our architecture is divided into multiple hierarchical levels, where the outermost level operates at full patch resolution, and each additional level operates at half of the spatial resolution per axis. For simplicity, we will first cover the cost at square resolutions of powers of two.

When designing the architecture for a specific resolution, we start with a dataset-dependent \textit{core} architecture, which, for natural images, typically includes one or two global-attention hierarchy levels that operate at $16^2$ or $16^2$ and $32^2$, respectively. Around that are a number of local attention levels. As this core only operates on a fixed resolution, it does not influence the asymptotic computational complexity of the overall model.

\oursubsubsection{Asymptotic Complexity Scaling}
When this architecture is adapted to a higher resolution, additional local attention levels with shared parameters are added to keep the innermost level operating at $16^2$. This means that the number of levels in our hierarchy scales with the number of image tokens as $\mathcal{O}(\log(n))$. While this might intuitively lead one to the conclusion of the overall complexity being $\mathcal{O}(n \log(n)d)$, as local attention layers' complexity is $\mathcal{O}(nd)$, the reduction in resolution at each level in the hierarchy has to be considered: due to the spatial downsampling, the number of tokens decreases by a factor of four at every level in the hierarchy, making the cost of the self-attention -- the only part of our model whose complexity does not scale linearly with token count -- of the additional levels
\[
    \sum_{l=1}^{\log_4(n) - \log_4(\resolution{core})} \frac{nd}{4^{l - 1}}.
\]
Factoring out $n$ and defining $m = l - 1$ yields
\[
    n \cdot \sum_{m=0}^{\log_4(n) - \log_4(\resolution{core}) - 1} d \cdot \left(\frac{1}{4}\right)^m,
\]
a (cut-off) geometric series with a common ratio of less than one, which means that, as the geometric series converges,
it does not affect the asymptotic complexity, making the cumulative complexity of the local self-attention of the additional levels $\mathcal{O}(nd)$.
Thus, as no other parts scale worse than $\mathcal{O}(nd)$ either, the overall complexity of the Hourglass Diffusion Transformer architecture, as the target resolution is increased, is $\mathcal{O}(nd)$.

\oursubsubsection{Local Complexity Scaling at Arbitrary Resolutions}
When the target resolution is increased by a factor smaller than a power of two per axis, the architecture is not adapted. This means that, for these intermediate resolutions, a different scaling behavior prevails. Here, the cost of the local attention levels, whose number does not change in this case, scales with $\mathcal{O}(nd)$ as before, but the global attention levels incur a quadratic increase in cost with the resolution. As the resolution is increased further, however, new levels are added, which reduce the resolution the global attention blocks operate at to their original values, and retaining the overall asymptotic scaling behavior of $\mathcal{O}(nd)$.

\begin{figure}[t]
    \centering
    \begin{adjustbox}{max width=\textwidth}
        \begin{tikzpicture}
    \datavisualization[
        scientific axes={
            clean,
            width=4cm,
            height=3.5cm,
        },
        x axis={
            logarithmic,
            label={Resolution (px)},
            ticks={
                none,
                at={128,256,512,1024},
                tick typesetter/.code={$\pgfmathprintnumber{#1}^2$},
            }
        },
        y axis={
            logarithmic,
            label={Computational Cost (GFLOP)},
            min value=10,
            max value=100000,
            ticks={
                none,
                at={10,100,1000,10000,100000},
            }
        },
        visualize as smooth line/.list={ditl2,adm,hditl2},
        ditl2={
            label in legend={text=Pixel-space DiT-B/4},
            style={
                mark=triangle*,
                visualizer color=ourorange,
                line width=0.3mm,
            },
        },
        adm={
            label in legend={text=ADM (parameter-matched)},
            style={
                mark=diamond*,
                visualizer color=ourblue,
                line width=0.3mm,
            },
        },
        hditl2={
            label in legend={text=\modelname-B/4 (\textbf{Ours})},
            style={
                mark=*,
                visualizer color=ourgreen,
                line width=0.3mm,
            },
        },
        legend={below, rows=3},
        style sheet=strong colors,
    ]
    
    data [set=ditl2, read from file=img/teaser/complexity_scaling_data/dit.csv, headline={x,y,r}]
    data [set=hditl2, read from file=img/teaser/complexity_scaling_data/hdit.csv, headline={x,y,r}]
    data [set=adm, read from file=img/teaser/complexity_scaling_data/adm.csv, headline={x,y,r}];

\end{tikzpicture}
    \end{adjustbox}
    \caption{Scaling of computational cost w.r.t. target resolution of our \modelname{}-B/4 model vs. DiT-B/4 \cite{peebles2023dit} and ADM \cite{dhariwal2021diffusion}.}
    \label{fig:complexity_scaling}
\end{figure}

\oursubsubsection{FLOP Comparison with DiT and Diffusion U-Nets}
While the asymptotic computational cost is important, it only describes how computational cost scales with resolution (in the theoretical limit). For practical purposes, it is important that the theoretical improvement from $\mathcal{O}(n^2d)$ to $\mathcal{O}(nd)$ from DiT to \modelname{} also results in lower FLOPs. To investigate this, we calculate the practical FLOPs for a parameter-matched pixel-space DiT and HDiT at various resolutions, which we show in \cref{fig:complexity_scaling}. We find that the theoretical improvements translate to real-world improvements, with HDiT already being more than 10 times more efficient at $256^2$ resolution, which further increases to a more than 100 times improvement for $1024^2$.
We also investigate a representative standard CNN-based diffusion U-Net \cite{dhariwal2021diffusion}. Here, we also find substantial performance gains of about 10 times at low resolutions, although the gap narrows at higher resolutions.

\section{Pixel-space vs. Latent Diffusion Models}\label{sec:pixel_latent_comparison}
{Extending upon the brief motivation presented in the introduction, we compare the advantages and disadvantages of pixel-space and latent \cite{rombach2022ldm} diffusion models.}

\subsection{Advantages of Pixel-space over Latent Diffusion}

{Several factors motivate the exploration of pixel-based alternatives:

1. Architectural Simplicity and Latent Space Limitations: Pixel-based models circumvent the need for complex latent space engineering, simplifying model architecture. Relying on a learned latent space introduces limitations tied to the VAE's representational capacity.

2. Quality Constraints and High-Frequency Information Loss: VAE-based diffusion models are inherently bounded by the reconstruction quality of the underlying VAE. Critically, VAEs are prone to losing high-frequency image details, hindering the generation of sharp and realistic images. We see evidence of this when roundtripping one of the generated images from our 557M ImageNet-$256^2$ model from \cref{fig:teaser} through the VAE used by DiT \cite{peebles2023dit} in \cref{fig:latent_quality_loss}

3. Fidelity Limitations for Image Manipulation: Faithful image reconstruction is crucial for downstream tasks like editing and transformation. VAEs often struggle with faithful reconstruction, limiting their applicability in these domains.

4. Challenges with Dynamic Thresholding and Intermediate Step Visualization: Integrating advanced sampling techniques like dynamic thresholding, as proposed in the DPM Solver \cite{lu2022dpm} literature, remains challenging within the latent space framework. Similarly, visualizing intermediate generation steps requires computationally expensive decoding, hindering iterative design processes.

5. Limited Compatibility with Classifier Guidance \cite{dhariwal2021diffusion}: Leveraging classifier guidance, a powerful technique for controlling image generation, proves difficult with latent space models. This difficulty arises from the mismatch between the pixel-space nature of most classifiers and the latent space representation of the diffusion model.

6. Empirical Evidence in Text-to-3D Synthesis: Recent work in text-to-3D generation has demonstrated superior performance with pixel-based diffusion models, highlighting their potential for high-fidelity synthesis \cite{shonenkov2023deepfloydif}.

7. Information Loss and Inpainting Challenges: The inherent information compression within the VAE latent space can negatively impact inpainting tasks. Specifically, it can lead to undesirable leakage of information from the surrounding regions into the inpainted area.}

\subsection{Advantages of Latent Diffusion Models}
Latent diffusion models \cite{rombach2022ldm} operate on the in the latent space of a variational auto-encoder. This allows for a substantial reduction in the spatial resolution, leading to a signficant computational reduction. The aforemnetioned reduction allows for usage of what would otherwise be computational infeasible choices, such as transformer models \cite{vaswani2017transformer}. The VAE inherently constrains the diffusion process to a manifold of plausible images. This effectively raises the lower bound on the average quality of generated images, leading to more consistent quality of images.

\section{Soft-Min-SNR Loss Weighting}\label{sec:min_soft_snr}
Min-SNR loss weighting \cite{hang2023minsnr} is a recently introduced training loss weighting scheme that improves diffusion model training. It adapts the SNR weighting scheme (for image data scaled to $\mathbf{x} \in \left[-1, 1\right]^{h \times w \times c}$)
\begin{equation}
    w_\text{SNR}(\sigma) = \frac{1}{\sigma^2}
\end{equation}
by clipping it at an SNR of $\gamma = 5$:
\begin{equation}
    w_\text{Min-SNR}(\sigma) = \min\left\{\frac{1}{\sigma^2}, \gamma\right\}.
\end{equation}
We utilize a slightly modified version that smoothes out the transition between the normal SNR weighting and the clipped section:
\begin{equation}
    w_\text{Soft-Min-SNR}(\sigma) = \frac{1}{\sigma^2 + \gamma^{-1}}.
\end{equation}
For $\sigma \ll \gamma$ and $\sigma \gg \gamma$, this matches Min-SNR, while providing a smooth transition
between both sections.

In practice, we also change the hyperparameter $\gamma$ from $\gamma = 5$ to $\gamma = 4$.

Plotting the resulting loss weight for both min-snr and our soft-min-snr as shown in \cref{fig:min_soft_snr_weighting} shows that our loss weighting is identical to min-snr, except for the transition, where it is significantly smoother. An ablation of our soft-min-snr compared to min-snr also shows that our loss weighting scheme leads to an improved FID score for our model, as shown in \cref{tab:additional_ablations}, steps \ablationidrope{} (SNR), \ablationidminsnrfive{} (Min-SNR, $\gamma = 5$), \ablationidminsnrfour{} (Min-SNR, $\gamma = 4$), \ablationidmss{} (Soft-Min-SNR, $\gamma = 4$).

\begin{figure}[H]
    \centering
    \begin{adjustbox}{max width=\columnwidth}
    \begin{minipage}[t]{\columnwidth}
        \centering
        \input{img/soft_min_snr_weight.pgf}
    \end{minipage}
    \end{adjustbox}
    \caption{The resulting loss weighting over $\sigma$ for our {\color{ourorange}soft-min-snr weighting} (orange) and {\color{ourblue}min-snr weighting} (blue) with $\gamma = 5$.}
    \label{fig:min_soft_snr_weighting}
\end{figure}

\section{Additional Experimental Results}
This section presents results for auxiliary experiments that provide additional context for the experiments presented in the main body of the paper.

\subsection{Additional Ablation Results}\label{sec:additional_ablations}
In \cref{tab:additional_ablations}, we present additional results for the main ablation study initially presented in \cref{sec:ablation_study}.

\oursubsubsection{Loss Weighting}
In \ablationidminsnrfive{} and \ablationidminsnrfour{}, we apply Min-SNR \cite{hang2023minsnr} loss weighting with the original hyperparameter $\gamma = 5$ and the value $\gamma = 4$ used for our Soft-Min-SNR. This shows that, in our setting, both the change of $\gamma$ and the smoother loss weighting help improve FID but that the smoothing plays a substantially larger role.

\oursubsubsection{Subtractive Ablations vs. DiT}
Extending our ablation in \cref{sec:ablation_study}, we also perform two subtractive ablations investigating the norm and activation choice in combination, whose results are shown in \cref{tab:additional_ablations}. Ablation step \ablationidadaln{} takes the full model but replaces the adaptive RMSNorm \cite{zhang2019root} with an adaptive layer norm \cite{ba2016layer} as used by DiT \cite{peebles2023dit}. Despite offering twice as many degrees of freedom due to predicting a shift in addition to the scale, we see no significant change in FID. Completely reverting to DiT-style blocks by changing GeGLU to GELU and adding an output gate controlled via the mapping network in step \ablationidadalngegluoutgate{} results in a worse FID, corroborating the results from the original ablation step \ablationidgeglu{}, even in combination with the different norm.

\oursubsubsection{Additional Baselines}
In \cref{sec:ablation_study}, we only present \ablationiddit{}, \ablationidditourblocks{}, and \ablationidditmss{} for simplicity. To evaluate the influence of our trainer and our loss weighting scheme, we also add an intermediate step, \ablationidditoursetup{}. This step wraps the official implementation of DiT-B/4 and adapts it to our codebase and trainer.\footnote{The pixel-space DiT \ablationidditoursetup{} was trained with an identical setup to the rest of our ablations except for the optimizer parameters: we initially tried training this model with our optimizer parameters but found it to both be unstable and worse than with the original parameters, so we used the original parameters from \cite{peebles2023dit} for the comparison.} This leads to a substantial reduction in FID compared to the original trainer, showing that it is important that the training setting matches the architecture. \ablationidditourblocks{} replaces the wrapped DiT model with a hyperparameter-matched single-level version of ablation step \ablationidbasic{}, matching the performance of the original DiT trained with the original codebase. On top of this setup, we also add soft-min-snr loss weighting to \ablationidditmss{} as in ablation step \ablationidmss{} to enable a fair comparison with our final model.

\newcommand{\ablationtablestep}[7]{{\color{gray}#1} & {\color{gray}#2} & {\color{gray}#3}}
\begin{table}[htb]
    \centering
    \caption{Additional ablation results on RGB ImageNet-$128^2$. Results already presented in \cref{tab:main_ablation} are presented in {\color{gray}gray font} as a reference.}
    \begin{adjustbox}{max width=\columnwidth}
        \begin{tabular}{l@{\hskip 1.5mm}|lc}\toprule
            \multicolumn{2}{l}{\textbf{Configuration}} & \textbf{FID}\lowerisbetter \\
            \midrule
            \multicolumn{3}{l}{\textbf{Baselines}}\\
            \ablationtablestep{\ablationiddit{} }{DiT-B/4 \cite{peebles2023dit}}{42.03}{106}{$\mathcal{O}(n^2)$}{657}{6,341}\\
            \ablationidditoursetup{} & \ablationiddit{} + Our Trainer & 69.86 \\
            \ablationtablestep{\ablationidditourblocks{} }{\ablationidditoursetup{} + Our Basic Blocks \& Mapping Network}{42.49}{106}{$\mathcal{O}(n^2)$}{657}{6,341}\\
            \midrule
            \ablationtablestep{\ablationidditmss{} }{\ablationidditourblocks{} + Soft-Min-SNR}{{30.71}}{106}{$\mathcal{O}(n^2)$}{657}{6,341}\\
            \midrule
            \midrule
            \multicolumn{3}{l}{\textbf{Ablation Steps}}\\
            \ablationtablestep{\ablationidbasic{} }{Global Attention Diffusion Hourglass (\cref{sec:hourglass_structure_diffusion})}{50.76}{{\color{white}0}32}{$\mathcal{O}(n^2)$}{114}{1,060}\\
            \ablationtablestep{\ablationidswi{} }{\ablationidbasic{} + Swin Attn. \cite{liu2021swin}}{55.93}{{\color{white}0}\textbf{29}}{\boldon}{{\color{white}0}60}{{\color{white}0,}185}\\
            \ablationtablestep{\ablationidnatten{} }{\ablationidbasic{} + Neighborhood Attn. \cite{hassani2023neighborhood}}{51.07}{{\color{white}0}\textbf{29}}{\boldon}{{\color{white}0}60}{{\color{white}0,}184}\\
            \ablationtablestep{\ablationidgeglu{} }{\ablationidnatten{} + GeGLU \cite{shazeer2020glu}}{44.36}{{\color{white}0}\underline{31}}{\boldon}{{\color{white}0}65}{{\color{white}0,}198}\\
            \ablationtablestep{\ablationidrope{} }{\ablationidgeglu{} + Axial RoPE (\cref{sec:block_design})}{41.41}{{\color{white}0}\underline{31}}{\boldon}{{\color{white}0}65}{{\color{white}0,}198}\\
            \midrule
            \ablationtablestep{\ablationidmss{} }{\ablationidrope{} + Soft-Min-SNR (\cref{sec:min_soft_snr})}{{27.74}}{{\color{white}0}\underline{31}}{\boldon}{{\color{white}0}65}{{\color{white}0,}198}\\
            \ablationidminsnrfive & \ablationidrope{} + Min-SNR \cite{hang2023minsnr} ($\gamma = 5$) & 36.65 \\
            \ablationidminsnrfour & \ablationidrope{} + Min-SNR \cite{hang2023minsnr} ($\gamma = 4$) & 35.62 \\
            \midrule
            \ablationidskipadd & \ablationidmss{} + Concatenation Skip & 33.75 \\
            \ablationidskipconcat & \ablationidmss{} + Additive Skip & {28.37} \\
            \midrule
            \ablationidadaln & \ablationidmss{} + AdaRMSNorm $\rightarrow$ AdaLN & 27.69 \\
            \ablationidadalngegluoutgate & \ablationidadaln{} + GeGLU $\rightarrow$ GeLU, DiT-style Output Gate & 30.66 \\
            \bottomrule
        \end{tabular}
    \end{adjustbox}
    \label{tab:additional_ablations}
\end{table}

\subsection{Effect of CFG for our 557M ImageNet-$256^2$ Model}
In addition to the analyses in \cref{sec:large_scale_imagenet}, which do not use classifier-free guidance (CFG) \cite{ho2021classifierfree}, we also analyze the FID-IS-tradeoff for difference guidance scales $w_{cfg}$ (we follow the guidance scale formulation used in \cite{saharia2022photorealistic}, where $w_{cfg} = 1$ corresponds to no classifier-free guidance being applied). The resulting curve is shown in \cref{fig:is_fid_imagenet}, with the lowest FID of $3.21$ being achieved around $w_{cfg} = 1.3$, with a corresponding IS of $220.6$.
\begin{figure}[H]
    \centering
    \begin{adjustbox}{max width=\columnwidth}
    \begin{minipage}[t]{\columnwidth}
        \centering
        \input{img/is_fid_plot.pgf}
    \end{minipage}
    \end{adjustbox}
    \caption{Inception Score vs. Fr\'echet Inception Distance at different classifier-free guidance weight scales (1 = no guidance) for our 557M ImageNet-$256^2$ model.}
    \label{fig:is_fid_imagenet}
\end{figure}

\subsection{Scaling Behavior}
To analyze the model's scaling behavior, we train a set of 9 models with varying model \& patch sizes. The hyperparameters are taken from our main 557M run and all models are trained in exactly the same setting for 1M steps. The shared hyperparameters are shown in \cref{tab:fig7_training_details_common}, and the individual run-specific details are shown in \cref{tab:fig7_training_details}. We show qualitative results in \cref{tab:fig7_eval}.

We show quantitative FID evaluations of the models in \cref{tab:fig7_eval}. Curiously, the most compute-intensive model (557M, patch size $4^2$) not only underperforms its smaller peers, but \textit{significantly} underperforms the identically-configured model from our large-scale ImageNet experiment (which, following a longer 2.2M steps of training, achieved an FID of 6.92 [3.21 with CFG]). We attribute this discrepancy to a suboptimal choice of hyperparameters, imposing a fixed learning rate (5e-4) and batch size (256) across all experiments. Our large-scale ImageNet experiment (\cref{sec:large_scale_imagenet}) mitigates this high learning rate by employing larger batch sizes later in training (see \cref{tab:training_details}). The notion of larger models' preferring larger batch sizes / lower learning rates, is corroborated by the line of work investigating $\mu$P-Parametrization \cite{yang2022feature,yang2022tensor}, which found that, using standard parameterizations (as we did for \modelname{}), a model's optimal learning rate decreases as size increases. Our learning rate of 5e-4 seems to work well for small models but seems to be too high for the larger models. Future work could change the parametrization to $\mu$P to enable using the same learning rate for all scales and revisit this experiment.

Qualitatively, we find that patch sizes as large as $16^2$ and $8^2$ are too ambitious for the transformer sizes ($\sim$100–500M) over which we ablated. Only the largest transformer (557M) achieved consistent coherence, and even then, only at the smallest patch size $4^2$. Studying the examplar sample grids in \cref{fig:scaling_visual}: we see that generation of round tennis balls or pumpkins succeeds at $4^2$ patch size for all transformer sizes, with some success also at $8^2$ patch size for the largest transformer. Balloons are coherent at $4^2$ patch size only, from the largest transformer and tenuously from the smallest. French loaves are coherent for the largest transformer only, at patch size $4^2$ (and tenuously $8^2$, notwithstanding questionable background forms), with texture best at $4^2$ (and arguably gummy at $8^2$). Ultimately, the $8^2$ patch size had too many coherence failures to recommend it, with even the largest transformer suffering discontinuous balloons, ill-defined cat eyelids, hyperbolic fox ears, vases with apertures, amorphous bullfrogs, wolf eye asymmetry, unbalanced poodles, and lemons eaten by their own leaves. Likewise, the medium transformer struggles at the lowest patch size $4^2$, exhibiting octopoid loaves, indistinct fox bodies, asymmetric cats and wolves, and fissured tennis balls. Coherence worsened further as model size decreased or as patch size increased.

\begin{table}[H]
    \centering
    \caption{Quantitative evaluation of our ImageNet-$256^2$ Transformer Size vs Patch Size sweep, illustrated in \cref{fig:scaling_visual}.}
    \begin{adjustbox}{max width=\textwidth}
    \begin{tabular}{lccc}
        \toprule
        \textbf{Parameter} & \textbf{Small} & \textbf{Medium} & \textbf{Large} \\
        \midrule
        \multicolumn{2}{l}{\textbf{Patch Size $16^2$}} \\
        Parameters & 116M & 267M & 507M \\
        FID\lowerisbetter & 90.6 & 51.8 & 37.6 \\
        \midrule
        \multicolumn{2}{l}{\textbf{Patch Size $8^2$}} \\
        Parameters & 134M & 294M & 547M \\
        FID\lowerisbetter & 50.3 & 30.9 & 33.6 \\
        \midrule
        \multicolumn{2}{l}{\textbf{Patch Size $4^4$}} \\
        Parameters & 139M & 302M & 557M \\
        FID\lowerisbetter & 21.6 & 24.0 & 29.3 \\
        \bottomrule
    \end{tabular}
    \end{adjustbox}
    \label{tab:fig7_eval}
\end{table}

\section{Implementation Details}
This section aims to answer potential questions about implementation details of \modelname{} for convenience. For further details, we refer to the reference implementation.

\subsection{Scaled Cosine Similarity Attention}\label{sec:cosine_sim_attn}
For the attention mechanism, we use a slight variation of the cosine similarity-based attention introduced in \cite{liu2022swin} they dub \textit{Scaled Cosine Attention} (a similar approach has also recently been used in \cite{karras2023edm2}): instead of computing the self-attention as
\begin{equation}
    \text{SA}(Q, K, V) = \text{softmax}\left(\frac{QK^\top}{\sqrt{d_\text{head}}}\right)V,
\end{equation}
they compute it as
\begin{equation}
    \text{SCA}(Q, K, V) = \text{softmax}\left(\frac{\text{sim}_\text{cos}(Q, K)}{\tau} + B_{ij}\right)V,
\end{equation}
with $\tau$ being a per-head per-layer learnable scalar, and $B_{ij}$ being the relative positional bias between pixel $i$ and $j$ (which we do not use in our models).
In practice, they parametrize $\tau$ based on a learnable parameter $\theta$ in the following way \cite{swin-v2-attention-implementation}:
\begin{equation}
    \frac{1}{\tau} = \exp\left(\min \left\{\theta, \log \frac{1}{0.01}\right\}\right),
\end{equation}
with $\theta$ being initialized to $\theta = \log 10$.

\oursubsubsection{Improving Scale Learning Stability}
We find that their parametrization of $\tau$ causes the learned scales to vary significantly during training, necessitating the clamping to a maximum value of 100 before exponentiation to prevent destabilization of the training. In this setting, we find that a significant number of scale factors $\tau$ reach this maximum value and values below 1 during our trainings. We speculate that this instability might be the cause of the behaviour observed in \cite{flash-cosine-sim-attention}, where using scaled cosine similarity attention was detrimental to the performance of generative models. To alleviate this problem, we find simply learning $\tau$ directly, as done for normal attention in \cite{henry2020qknorm}, prevents this large variance of its values in our models, with our converged models' scale typically reaching a range between 5 and 50.

\subsection{Axial RoPE}\label{sec:axial_rope_details}
We extend rotary positional embeddings \cite{su2022roformer} to 2D image data.
We split the encoding to operate independently along each axis, applying RoPE for each spatial axis to half of the query and key each.
Empirically, we find that applying this embedding scheme to only half of key \& query and leaving the other half unmodified (see \cref{fig:axial_rope} for an illustration) results in better performance than applying it for the full key \& query.

\begin{figure}[htb]
    \centering
    \begin{adjustbox}{max width=\textwidth}
        \newcommand{\ariblockheight}{7mm}
\newcommand{\ariblockquarterwidth}{14mm}
\newcommand{\ariblockhalfwidth}{28mm}
\newcommand{\ariblockwidth}{56mm}
\begin{tikzpicture}
    \node[standard node={blue},dashed,fill=none,minimum width=\ariblockquarterwidth, minimum height=\ariblockheight] (h) {\color{ourblue}$\mathbf{R}_y$};
    \node[standard node={green},dashed,fill=none,minimum width=\ariblockquarterwidth, minimum height=\ariblockheight,right=0mm of h] (w) {\color{ourgreen}$\mathbf{R}_x$};
    \node[standard node={gray},dashed,fill=none,minimum width=\ariblockhalfwidth, minimum height=\ariblockheight,right=0mm of w] (empty) {\color{ourgray}1};
    \begin{scope}[on background layer]
        \node[
            standard node={white},
            fit=(h) (h) (h) (empty),
            inner sep=.2mm,
        ] (full) {};
    \end{scope}
    \draw[decorate,decoration={calligraphic brace},line width=0.04cm] ($(full.east) + (0,-5mm)$) -- node[below] {$d_\text{head}$} ($(full.west) + (0,-5mm)$);
\end{tikzpicture}
    \end{adjustbox}
    \caption{Illustration of our 2D axial RoPE embedding scheme. The rotation for the vertical position $\mathbf{R}_y$ and horizontal position $\mathbf{R}_x$ are applied to one quarter of the key/query each, while the rest is left unaffected.}
    \label{fig:axial_rope}
\end{figure}

\subsection{Conditioning}
\oursubsubsection{Adaptive RMSNorm}
Following common practice, we implement conditioning using adaptive norms \cite{huang2017adain}, where we apply a standard RMSNorm \cite{zhang2019root}
\begin{equation}
    x_{i,\mathrm{scaled}} = \frac{x_i}{\mathrm{RMS}(\mathbf{x})} \cdot g_i, \text{with } \mathrm{RMS}(\mathbf{x}) = \sqrt{\frac{1}{N}\sum_{i = 1}^{N} x_i^2},
\end{equation}
with $\mathbf{g}$ being predicted from the mapping network based on the conditioning $\mathbf{c}$ instead of being a learned vector as $\mathbf{g} = 1 + \mathrm{mapping}(\mathbf{c})$. At initialization, the final linear projection is initialized to zero, making 

\oursubsubsection{Mapping Network}
The prediction of the RMSNorm scales is implemented via a mapping network that takes the diffusion timestep, the class conditioning, and, optionally, augmentation information to prevent augmentation leakage \cite{karras2020styleganada}.

The mapping network consists of $N$ blocks that process the conditioning information. The blocks' architecture is almost identical to our pointwise FFN block (see \cref{fig:ffn_diagram}). For the initial embedding, we use a standard learnable embedding for the class conditioning, and random fourier features (following \cite{karras2022elucidating}) followed by linear projections for the diffusion timestep and augmentation conditioning. An overview of the network and block structure is given in \cref{fig:mapping_diagram}.

\begin{figure}[htb]
    \centering
    \begin{subfigure}[h]{.5\columnwidth}
        \centering
        \scalebox{.65}{%
            \begin{tikzpicture}
    \node[standard node={gray},minimum width=1.5cm] (inputtime) {Diff.\\Time};
    \draw[standard arrow] ($(inputtime.south) + (0,-.4cm)$) -- (inputtime);
    \node[standard node={gray},dashed,minimum width=1.5cm,left=of inputtime] (inputcond) {Class\\Cond.};
    \draw[standard arrow] ($(inputcond.south) + (0,-.4cm)$) -- (inputcond);
    \node[standard node={gray},dashed,minimum width=1.5cm,right=of inputtime] (inputaugcond) {Aug.\\Cond.};
    \draw[standard arrow] ($(inputaugcond.south) + (0,-.4cm)$) -- (inputaugcond);

    \node[standard node={blue},minimum width=1.5cm, above=of inputtime] (timeemb) {Rand.\\Fourier};
    \draw[standard arrow] (inputtime) -- (timeemb);
    \node[standard node={green},minimum width=1.5cm, above=of timeemb] (timeproj) {Linear};
    \draw[standard arrow] (timeemb) -- (timeproj);

    \node[standard node={green},minimum width=1.5cm, left=of timeproj] (classemb) {Emb.};
    \draw[standard arrow] (inputcond) -- (classemb);

    \node[standard node={blue},minimum width=1.5cm, above=of inputaugcond] (augemb) {Rand.\\Fourier};
    \draw[standard arrow] (inputaugcond) -- (augemb);
    \node[standard node={green},minimum width=1.5cm, above=of augemb] (augproj) {Linear};
    \draw[standard arrow] (augemb) -- (augproj);
    
    \node[standard node={blue}, above=of timeproj] (inputnorm) {RMSNorm};
    \draw[standard arrow] (timeproj) -- (inputnorm);
    \draw[standard arrow] (classemb.north) -- ++(0,0.15cm) -| (inputnorm);
    \draw[standard arrow] (augproj.north) -- ++(0,0.15cm) -| (inputnorm);
    
    \node[standard node={yellow}, above=of inputnorm] (blocks1) {Mapping Block};
    \draw[standard arrow] (inputnorm) -- (blocks1);

    \node[above=of blocks1] (blocks) {...};
    \draw[standard arrow] (blocks1) -- (blocks);
    
    \node[standard node={yellow}, above=of blocks] (blocksN) {Mapping Block};
    \draw[standard arrow] (blocks) -- (blocksN);

    \draw [decorate,decoration={calligraphic brace},line width=0.04cm] ($(blocksN.east) + (1mm,3mm)$) -- node[right] {$N$} ($(blocks1.east) + (1mm,-3mm)$);

    \node[standard node={blue}, above=of blocksN] (outputnorm) {RMSNorm};
    \draw[standard arrow] (blocksN) -- (outputnorm);
    
    \draw[standard arrow] (outputnorm.north) -- ++(0,0.4cm);

    \begin{scope}[on background layer]
        \node[
            standard node={white},
            fit=(outputnorm) (inputcond) (inputcond) (inputaugcond)
        ] {};
    \end{scope}
\end{tikzpicture}
        }
        \caption{\modelname{} mapping network.}
    \end{subfigure}%
    ~
    \begin{subfigure}[h]{.5\columnwidth}
        \centering
        \scalebox{.65}{%
            \begin{tikzpicture}
    \node[standard node={gray}] (inputtokens) {Input};
    \draw[standard arrow] ($(inputtokens.south) + (0,-.4cm)$) -- (inputtokens);

    \node[standard node={blue},above=of inputtokens] (innorm) {RMSNorm};
    \draw[standard arrow] (inputtokens) -- (innorm);

    \node[above=of innorm,minimum width=0cm,minimum height=9mm] (linearcenter) {};
    \node[standard node={green},minimum width=1cm,left=of linearcenter.east] (linear1) {Linear};
    \node[standard node={green},minimum width=1cm,right=of linearcenter.west] (linear2) {Linear};
    \draw[standard arrow] (innorm.north) -- ++(0,0.15cm) -| (linear2);
    \draw[standard arrow] (innorm.north) -- ++(0,0.15cm) -| (linear1);
    
    \node[standard node={blue},minimum width=1cm,above=of linear1] (gelu) {GELU};
    \draw[standard arrow] (linear1) -- (gelu);
    \node[above=of linearcenter,minimum width=0cm,minimum height=3mm] (linearcenter2) {};
    
    \node[standard node circle={white}, above=of linearcenter2] (prod) {$\odot$};
    \draw[standard arrow] (gelu) |- (prod);
    \draw[standard arrow] (linear2) |- (prod);
    \node[standard node={blue},dashed, above=of prod] (dropout) {Dropout};
    \draw[standard arrow] (prod) -- (dropout);
    \node[standard node={green}, above=of dropout] (downproj) {Linear};
    \draw[standard arrow] (dropout) -- (downproj);
    \node[standard node circle={white}, above=of downproj] (ffnskip) {+};
    \draw[standard arrow] (downproj) -- (ffnskip);
    \draw[standard arrow] (inputtokens.north) -- ++(0,0.15cm) -| ++(-1.9cm,0) |- ($(ffnskip.west) + (-.1cm,0)$) node [midway] (skiparrowmidway) {};
    
    \draw[standard arrow] (ffnskip.north) -- ++(0,0.4cm);

    \node[right=of linear2] (rightspacing) {};
    \begin{scope}[on background layer] 
        \node[
            standard node={white},
            fit=(ffnskip) (skiparrowmidway) (rightspacing) (inputtokens)
        ] {};
    \end{scope}
    
    \begin{scope}[on background layer] 
        \node[
            standard node={white},
            label={[rotate=90,xshift=-1.85cm,yshift=-2.5mm]right:{GEGLU \cite{shazeer2020glu}}},
            fit=(prod) (linear1) (linear1) (linear2)
        ] {};
    \end{scope}
\end{tikzpicture}
        }
        \caption{\modelname{} mapping block.}
    \end{subfigure}
    \caption{An overview of our mapping network architecture.}
    \label{fig:mapping_diagram}
\end{figure}

This general conditioning embedding is then passed to each block in the main network, where it is projected locally to obtain the relevant information for that block and obtain the feature scales.

\subsection{Token Merging \& Splitting}
For token merging and splitting inside our architecture, we follow a standard Pixel-Shuffle~\cite{shi2016pixelshuffle}-based approach. Token merging is implemented as a reshaping of the tensor from $B \times H \times W \times C$ to $B \times \frac{H}{M} \times \frac{W}{M} \times CM^2$ (\textit{Pixel-UnShuffle}), with $M = 2$, followed by a pointwise linear projection to adjust the channel count to the appropriate model width at that level.
Similarly, token splitting is implemented as a pointwise linear projection, bringing the channel count from the model width to $CM^2$, followed by a reshaping $B \times H \times W \times CM^2$ to $B \times HM \times WM \times C$ (\textit{Pixel-Shuffle}).
This follows various previous implementations such as \cite{zamir2022restormer}.

\onecolumn
\section{Experiment Details}
We provide an overview of all relevant hyperparameters, training hardware, and time for the experiments presented in this paper in \cref{tab:training_details} and \cref{tab:fig7_training_details_all}.

\addtocounter{footnote}{1}\addtocounter{Hfootnote}{1}\footnotetext{\label{footnote:ablationtrainingdetails}The other ablation steps generally use the same parameters, except for the architectural changes indicated in the experiment description.}
\addtocounter{footnote}{1}\addtocounter{Hfootnote}{1}\footnotetext{\label{footnote:imagenet557msteps}We initially trained for 2M steps. We then experimented with progressively increasing the batch size (waiting until the loss plateaued to a new, lower level each time), training at batch size 512 for an additional 50k steps, at batch size 1024 for 100k, and at batch size 2048 for 50k steps.}
\addtocounter{footnote}{1}\addtocounter{Hfootnote}{1}\footnotetext{\label{footnote:wallclocktime}Wall clock time, including startup, validation, checkpoint saving, etc.}

\begin{table}[H]
    \centering
    \caption{Details of our training and inference setup.}
    \begin{adjustbox}{max width=\textwidth}
    \begin{tabular}{lccc}
        \toprule
        \textbf{Parameter} & \textbf{ImageNet}-$128^2$ & \textbf{FFHQ}-$1024^2$ & \textbf{ImageNet}-$256^2$ \\
        \midrule
        Experiment & Ablation \ablationidmss{}\textsuperscript{\ref{footnote:ablationtrainingdetails}} (\cref{sec:ablation_study}) & High-Res Synthesis (\cref{sec:high_res_image_synthesis}) & Large-Scale (\cref{sec:large_scale_imagenet}) \\
        Parameters & 117M & 85M & 557M \\
        GFLOP/forward & 31 & 206 & 198 \\
        \arrayrulecolor{black!50} \midrule \arrayrulecolor{black}
        Training Steps & 400k & 1M & 2.2M \\
        Batch Size & 256 & 256 & 256+\textsuperscript{\ref{footnote:imagenet557msteps}} \\
        Precision & bfloat16 & bfloat16 & bfloat16 \\
        Training Hardware & 4 A100 80GiB & 64 A100 80GiB & 8 H100 80GiB \\
        Training Time & 15 hours\textsuperscript{\ref{footnote:wallclocktime}} & 5 days\textsuperscript{\ref{footnote:wallclocktime}} & 7.6 days \\
        \midrule
        Patch Size & 4 & 4 & 4 \\
        Levels (Local + Global Attention) & 1 + 1 & 3 + 2 & 2 + 1 \\
        Depth & [2, 11] & [2, 2, 2, 2, 2] & [2, 2, 16] \\
        Widths & [384, 768] & [128, 256, 384, 768, 1024] & [384, 768, 1536] \\
        Attention Heads (Width / Head Dim) & [6, 12] & [2, 4, 6, 12, 16] & [6, 12, 24] \\
        Attention Head Dim & 64 & 64 & 64 \\
        Neighborhood Kernel Size & 7 & 7 & 7 \\
        \arrayrulecolor{black!50} \midrule \arrayrulecolor{black}
        Mapping Depth & 1 & 2 & 2 \\
        Mapping Width & 768 & 768 & 768 \\
        \midrule
        Data Sigma & 0.5 & 0.5 & 0.5 \\
        Sigma Range & [1e-3, 1e3] & [1e-3, 1e3] & [1e-3, 1e3] \\
        Sigma Sampling Density & interpolated cosine & interpolated cosine & interpolated cosine \\
        \arrayrulecolor{black!50} \midrule \arrayrulecolor{black}
        Augmentation Probability & 0 & 0.12 & 0 \\
        Dropout Rate & 0 & [0, 0, 0, 0, 0.1] & 0 \\
        Conditioning Dropout Rate & 0.1 & 0.1 & 0.1 \\
        \arrayrulecolor{black!50} \midrule \arrayrulecolor{black}
        Optimizer & AdamW & AdamW & AdamW \\
        Learning Rate & 5e-4 & 5e-4 & 5e-4 \\
        Betas & [0.9, 0.95] & [0.9, 0.95] & [0.9, 0.95] \\
        Eps & 1e-8 & 1e-8 & 1e-8 \\
        Weight Decay & 1e-2 & 1e-2 & 1e-2 \\
        \arrayrulecolor{black!50} \midrule \arrayrulecolor{black}
        EMA Decay & 0.9999 & 0.9999 & 0.9999 \\
        \midrule
        Sampler & DPM++(3M) SDE & DPM++(3M) SDE & DPM++(3M) SDE \\
        Sampling Steps & 50 & 50 & 50 \\
        \bottomrule
    \end{tabular}
    \end{adjustbox}
    \label{tab:training_details}
    \vspace{25mm}
\end{table}

\begin{table}[H]
    \centering
    \caption{Details of our ImageNet-$256^2$ Transformer Size vs. Patch Size training and inference setup.}
    \label{tab:fig7_training_details_all}
    \begin{subtable}[h]{.35\columnwidth}
        \centering
        \caption{Details common to all configs in our Transformer Size vs. Patch Size experiments.}
        \begin{adjustbox}{max width=\textwidth}
        \begin{tabular}{lccc}
            \toprule
            \textbf{Parameter} \\
            \midrule
            Training Steps & 1M \\
            Batch Size & 256 \\
            Precision & bfloat16 \\
            \midrule
            Attention Head Dim & 64 \\
            Neighborhood Kernel Size\textsuperscript{\ref{footnote:fig7_neighbourhood_size}} & 7 \\
            \arrayrulecolor{black!50} \midrule \arrayrulecolor{black}
            Mapping Depth & 2 \\
            Mapping Width & 768 \\
            \midrule
            Data Sigma & 0.5 \\
            Sigma Range & [1e-3, 1e3] \\
            Sigma Sampling Density & interpolated cosine \\
            \arrayrulecolor{black!50} \midrule \arrayrulecolor{black}
            Augmentation Probability & 0 \\
            Dropout Rate & 0 \\
            Conditioning Dropout Rate & 0.1 \\
            \arrayrulecolor{black!50} \midrule \arrayrulecolor{black}
            Optimizer & AdamW \\
            Learning Rate & 5e-4 \\
            Betas & [0.9, 0.95] \\
            Eps & 1e-8 & \\
            Weight Decay & 1e-2 \\
            \arrayrulecolor{black!50} \midrule \arrayrulecolor{black}
            EMA Decay & 0.9999 \\
            \midrule
            Sampler & DPM++(2M) SDE \\
            Sampling Steps & 50 \\
            \bottomrule
        \end{tabular}
        \end{adjustbox}
        \label{tab:fig7_training_details_common}
    \end{subtable}%
    ~
    \begin{subtable}[h]{.65\columnwidth}
        \centering
        \caption{Config-specific details of our Transformer Size vs. Patch Size experiments.}
        \begin{adjustbox}{max width=\textwidth}
        \begin{tabular}{lccc}
            \toprule
            \textbf{Parameter} & \textbf{Small} & \textbf{Medium} & \textbf{Large} \\
            \midrule
            \textbf{Patch Size $16^2$} \\
            Parameters & 116M & 267M & 507M \\
            GFLOP/forward & 29 & 68 & 129 \\
            Training Hardware & 4 A100 80GiB & 4 A100 40GiB & 4 A100 40GiB \\
            Training Time\textsuperscript{\ref{footnote:wallclocktime}} & 1.1 days & 2.5 days & 4.4 days \\
            \arrayrulecolor{black!50} \midrule \arrayrulecolor{black}
            Levels (Local + Global Attention) & 0 + 1 & 0 + 1 & 0 + 1 \\
            Depth & 8 & 12 & 16 \\
            Widths & 1024 & 1280 & 1536 \\
            Attention Heads (Width / Head Dim) & 16 & 20 & 24 \\
            \midrule
            \textbf{Patch Size $8^2$} \\
            Parameters & 134M & 294M & 547M \\
            GFLOP/forward & 44 & 91 & 163 \\
            Training Hardware & 4 A100 80GiB & 2$\times$4 A100 40GiB & 2$\times$4 A100 40GiB \\
            Training Time\textsuperscript{\ref{footnote:wallclocktime}} & 2.6 days & 2.2 days & 3.6 days \\
            \arrayrulecolor{black!50} \midrule \arrayrulecolor{black}
            Levels (Local + Global Attention) & 1 + 1 & 1 + 1 & 1 + 1 \\
            Depth & [2, 8] & [2, 12] & [2, 16] \\
            Widths & [512, 1024] & [640, 1280] & [768, 1536] \\
            Attention Heads (Width / Head Dim) & [8, 16] & [10, 20] & [12, 24] \\
            \midrule
            \textbf{Patch Size $4^4$} \\
            Parameters & 139M & 302M & 557M \\
            GFLOP/forward & 60 & 115 & 198 \\
            Training Hardware & 4xA100 40GiB & 2x4xA100 40GiB & 2x4xA100 40GiB \\
            Training Time\textsuperscript{\ref{footnote:wallclocktime}} & 3.7 days & 3.3 days & 6.1 days \\
            \arrayrulecolor{black!50} \midrule \arrayrulecolor{black}
            Levels (Local + Global Attention) & 2 + 1 & 2 + 1 & 2 + 1 \\
            Depth & [2, 2, 8] & [2, 2, 12] & [2, 2, 16] \\
            Widths & [256, 512, 1024] & [320, 640, 1280] & [384, 768, 1536] \\
            Attention Heads (Width / Head Dim) & [4, 8, 16] & [5, 10, 20] & [6, 12, 24] \\
            \bottomrule
        \end{tabular}
        \end{adjustbox}
        \label{tab:fig7_training_details}
    \end{subtable}
\end{table}

\addtocounter{footnote}{1}\interfootnotelinepenalty=10000\footnotetext{\label{footnote:fig7_neighbourhood_size}Transformers with patch size $16^2$ did not possess any neighborhood attention levels}

\clearpage
\section{Scaling Samples}
We provide an equivalent of Fig. 7 from DiT \cite{peebles2023dit}, where samples are generated with fixed random seed across multiple patch sizes and transformer scales, in \cref{fig:scaling_visual}. The quality of generated samples increases with smaller patch sizes and larger transformers, matching the findings for DiT and demonstrating the scalability of \modelname{}.

\begin{figure}[H]
    \centering
    \adjustbox{max width=\columnwidth}{
    \input{img/scaling_visual}
    }
    \caption{Scaling behaviour of our \modelname{} across different model and patch sizes on pixel-space ImageNet-$256^2$. All models used to generate samples for this figure have been trained for 1M steps, and samples have been generated without classifier-free guidance. Patch sizes shown are $\{16, 8, 4\}$, transformer sizes approximately double at each step, up to our 557M ImageNet-$256^2$ model. See \cref{tab:fig7_training_details_all} for detailed hyperparameters.}
    \label{fig:scaling_visual}
\end{figure}

\section{Our FFHQ-$1024^2$ Samples}

\begin{figure}[H]
    \centering
    \includegraphics[width=\linewidth]{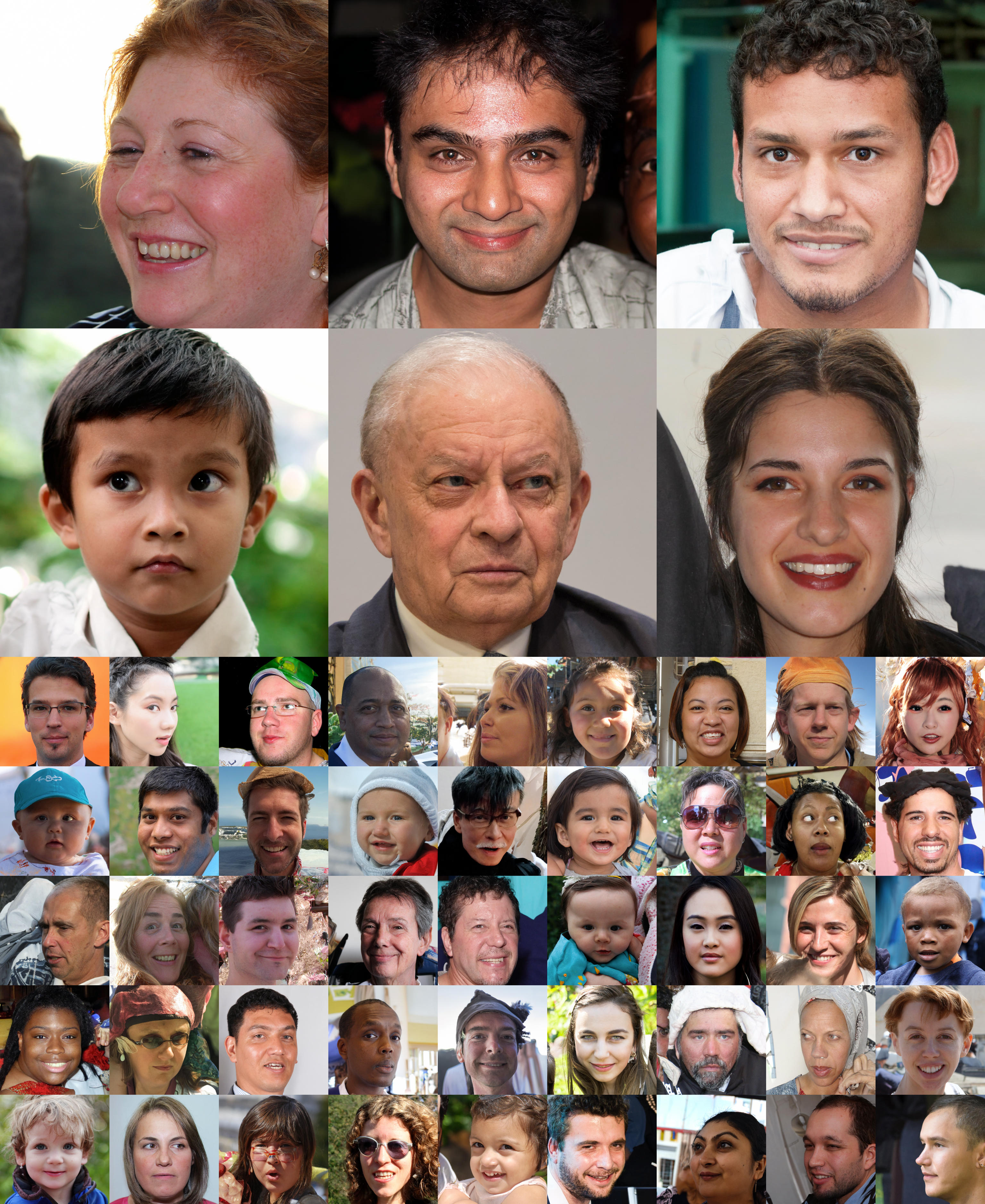}
    \caption{\textbf{Uncurated} samples from our 85M \modelname{} FFHQ-$1024^2$ model.}
    \label{fig:ffhq_uncurated}
\end{figure}

\section{FFHQ-$1024^2$ Reference Samples}

\begin{figure}[H]
    \centering
    \begin{tabular}{c}\textbf{\modelname{}}\end{tabular}
    \includegraphics[width=\linewidth]{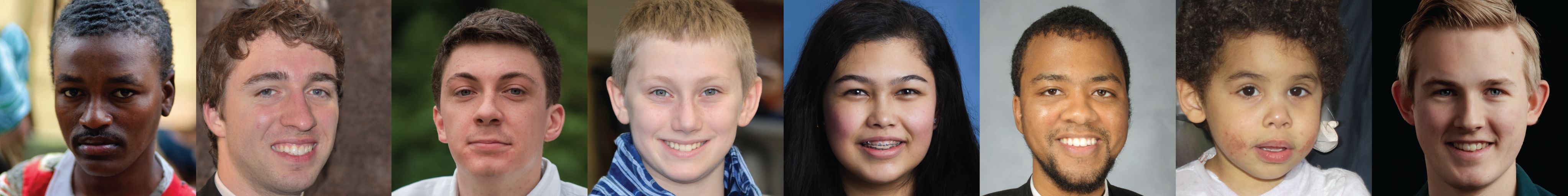}
    \begin{tabular}{c}\textbf{StyleGAN2 \cite{karras2019stylegan2}}\end{tabular}
    \includegraphics[width=\linewidth]{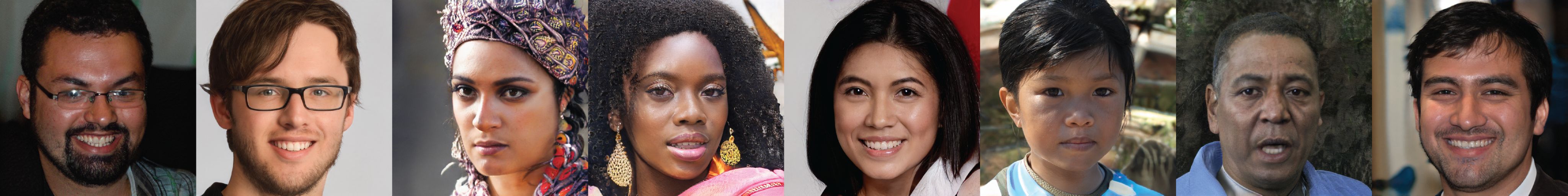}
    \begin{tabular}{c}\textbf{StyleGAN3-R \cite{karras2021stylegan3}}\end{tabular}
    \includegraphics[width=\linewidth]{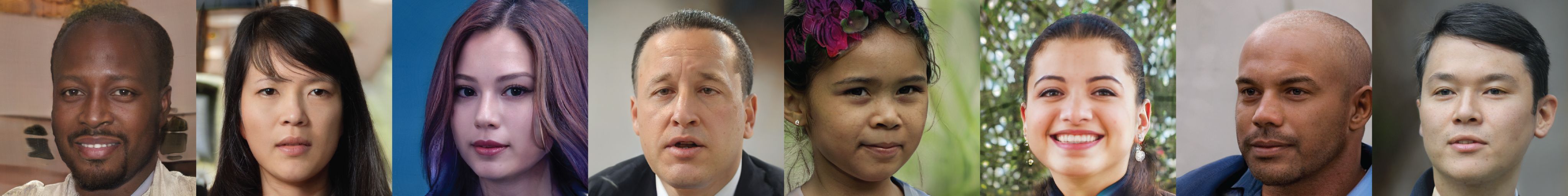}
    \begin{tabular}{c}\textbf{StyleGAN3-T \cite{karras2021stylegan3}}\end{tabular}
    \includegraphics[width=\linewidth]{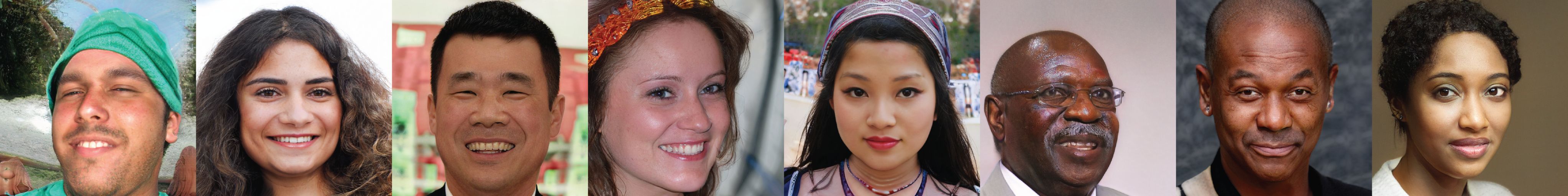}
    \begin{tabular}{c}\textbf{StyleSwin \cite{zhang2022styleswin}}\end{tabular}
    \includegraphics[width=\linewidth]{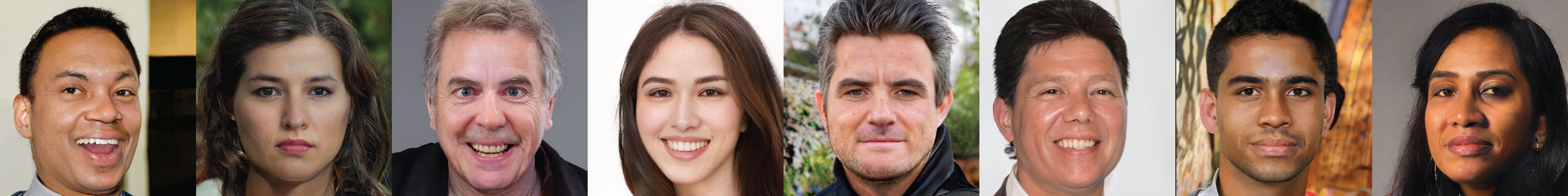}
    \begin{tabular}{c}\textbf{StyleGAN-XL \cite{sauer2022styleganxl}}\end{tabular}
    \includegraphics[width=\linewidth]{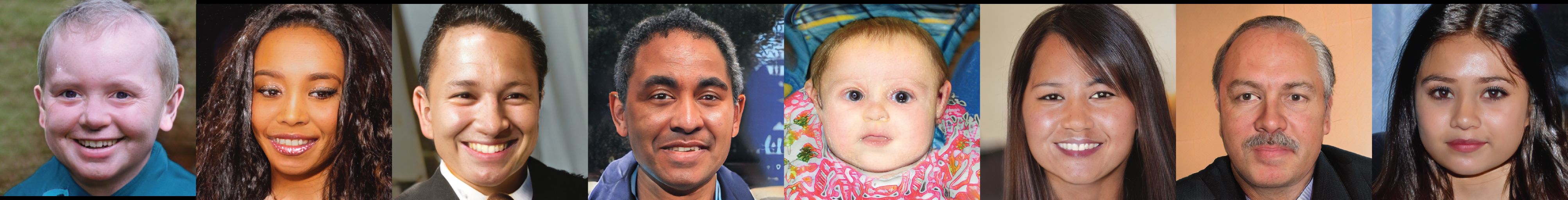}
    \begin{tabular}{c}\textbf{NCSN++ \cite{song2021scorebased}}\end{tabular}
    \includegraphics[width=\linewidth]{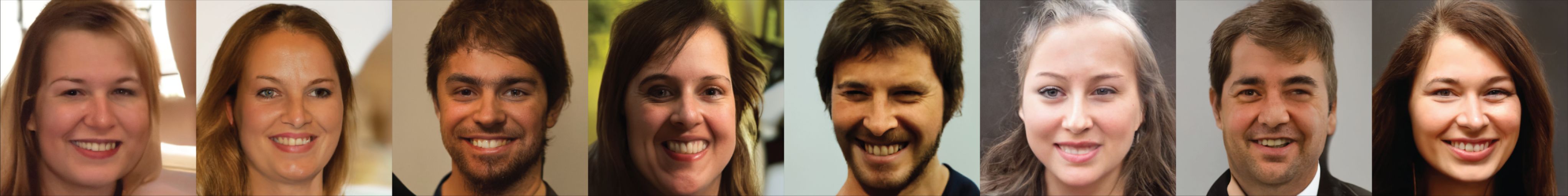}
    \caption{\textbf{Curated} FFHQ-$1024^2$ reference samples from Hourglass, StyleGAN2 \cite{karras2019stylegan2}, StyleGAN3-R \cite{karras2021stylegan3}, StyleGAN3-T \cite{karras2021stylegan3}, StyleSwin \cite{zhang2022styleswin}, StyleGAN-XL \cite{sauer2022styleganxl}, and NCSN++ \cite{song2021scorebased} models.}
    \label{fig:ffhq_curated}
\end{figure}

\section{Our ImageNet-$256^2$ Samples}

\begin{figure}[H]
    \centering
    \includegraphics[width=\linewidth]{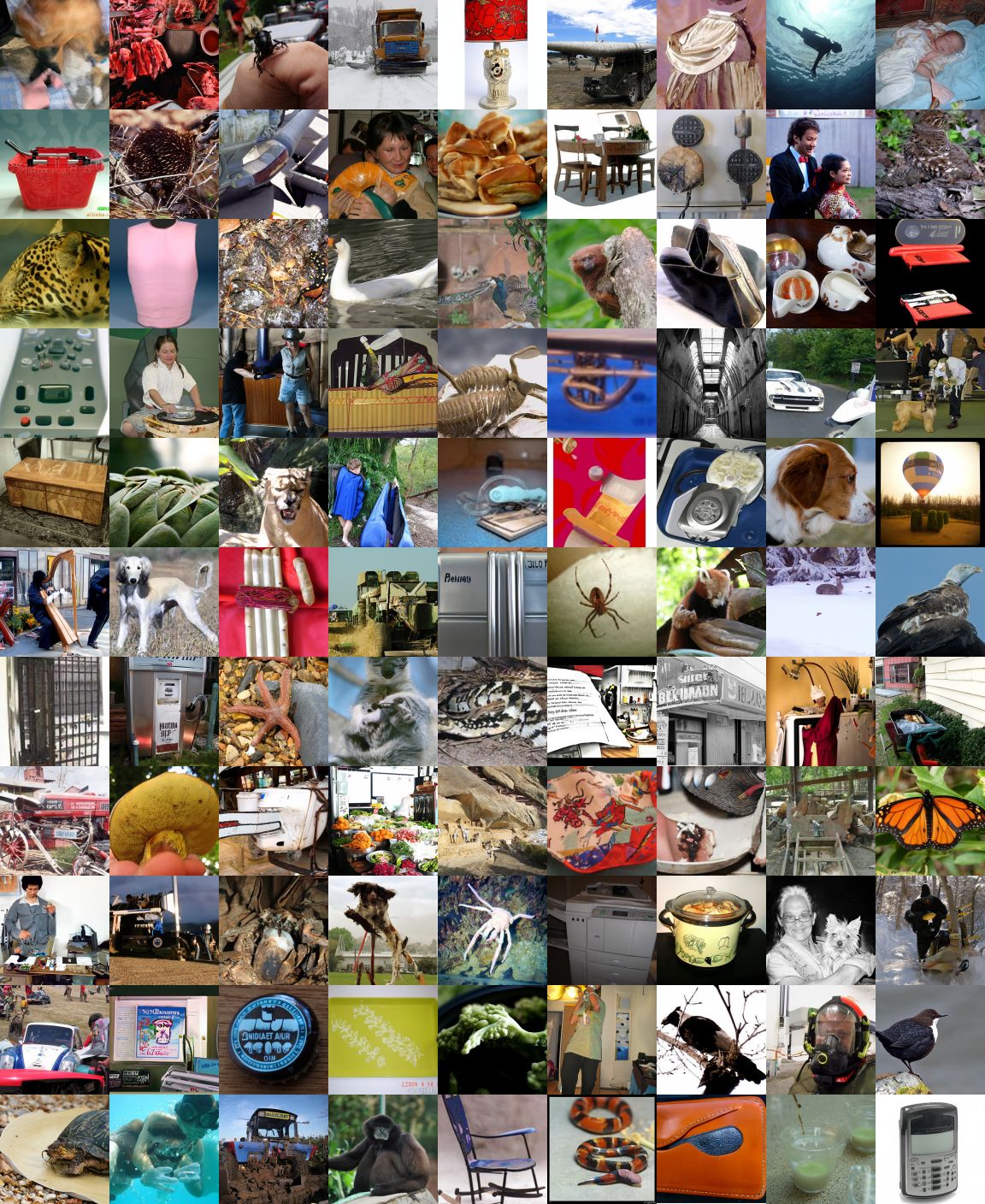}
    \caption{\textbf{Uncurated} random class-conditional samples from our 557M \modelname{} ImageNet-$256^2$ model.}
    \label{fig:imagenet_uncurated}
\end{figure}

\begin{figure}[H]
    \centering
    \includegraphics[width=\linewidth]{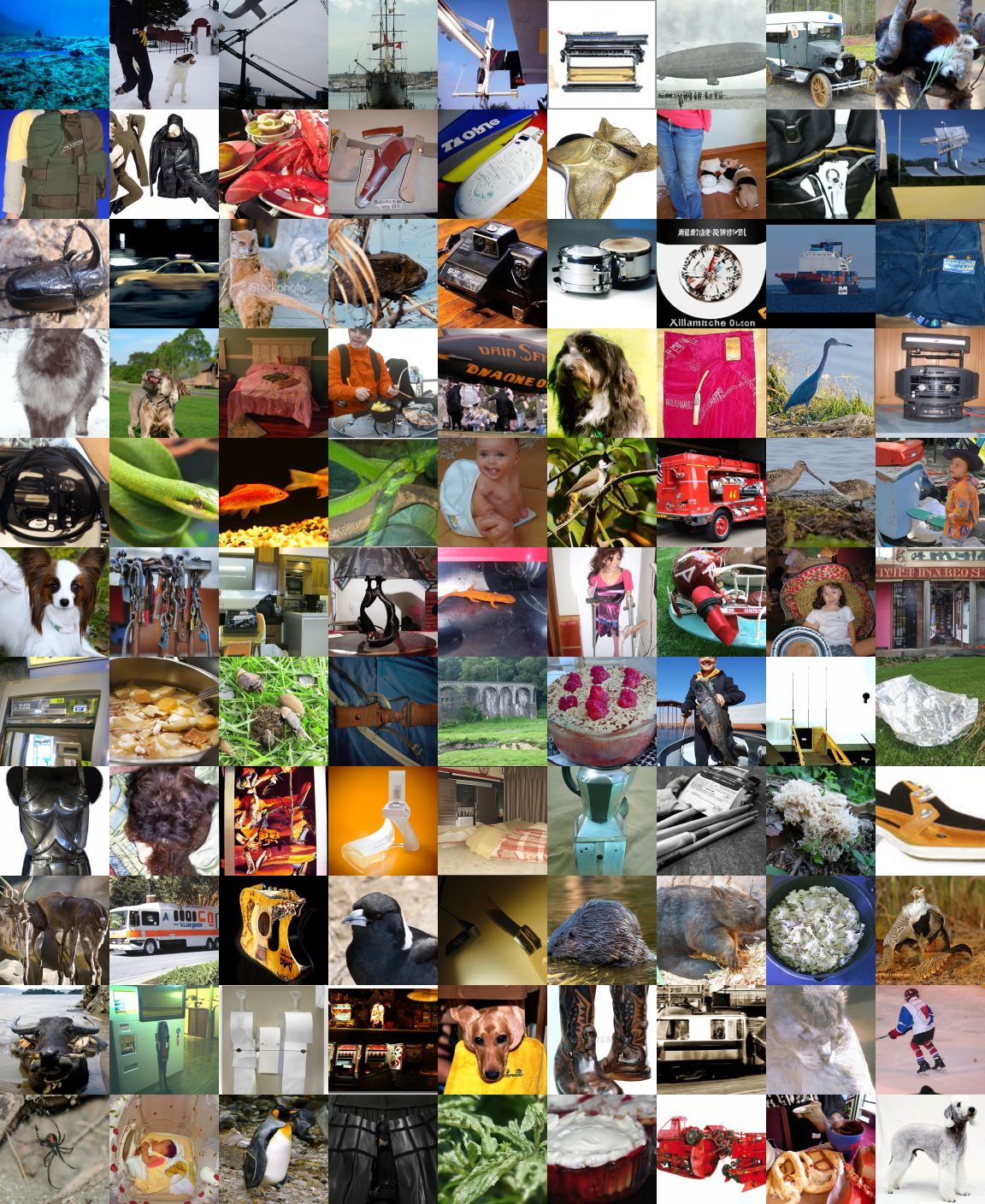}
    \caption{More \textbf{uncurated} random class-conditional samples from our \modelname{}-557M ImageNet-$256^2$ model.}
    \label{fig:imagenet_uncurated_2}
\end{figure}

\clearpage

\end{document}